\documentclass[letterpaper]{article} 
\usepackage{aaai2026}  
\usepackage{times}  
\usepackage{helvet}  
\usepackage{courier}  
\usepackage[hyphens]{url}  
\usepackage{graphicx} 
\usepackage{natbib}  
\usepackage{caption} 
\usepackage{algorithm}
\usepackage{algpseudocode}

\usepackage{multirow}
\usepackage{bbding}
\usepackage{amsmath}
\usepackage{cleveref}
\usepackage{colortbl}
\usepackage{booktabs}
\usepackage{xcolor}
\usepackage{amssymb}

\usepackage{newfloat}
\usepackage{listings}
\DeclareCaptionStyle{ruled}{labelfont=normalfont,labelsep=colon,strut=off} 
\floatstyle{ruled}
\newfloat{listing}{tb}{lst}{}
\floatname{listing}{Listing}
\title{DiffRefiner: Coarse to Fine Trajectory Planning via Diffusion Refinement with Semantic Interaction for End to End Autonomous Driving}
\author{
    Liuhan Yin\textsuperscript{\rm 1,2}\equalcontrib\thanks{Work done during an internship at Nullmax.},
    Runkun Ju\textsuperscript{\rm 2}\equalcontrib,
    Guodong Guo\textsuperscript{\rm 2},
    Erkang Cheng\textsuperscript{\rm 2}\thanks{Corresponding author.}
}
\affiliations{
    \textsuperscript{\rm 1}Polytechnic Institute, Zhejiang University, China\\
    \textsuperscript{\rm 2}Nullmax, China\\

    liuhanyin@zju.edu.cn, \{jurunkun, guoguodong, chengerkang\}@nullmax.ai
}

\usepackage{bibentry}

\begin{document}

\maketitle

\begin{abstract}
Unlike discriminative approaches in autonomous driving that predict a fixed set of candidate trajectories of the ego vehicle, generative methods, such as diffusion models, learn the underlying distribution of future motion, enabling more flexible trajectory prediction. 
However, since these methods typically rely on denoising human-crafted trajectory anchors or random noise, there remains significant room for improvement.
In this paper, we propose DiffRefiner, a novel two-stage trajectory prediction framework. 
The first stage uses a transformer-based \textit{Proposal Decoder} to generate coarse trajectory predictions by regressing from sensor inputs using predefined trajectory anchors. The second stage applies a \textit{Diffusion Refiner} that iteratively denoises and refines these initial predictions. 
In this way, we enhance the performance of diffusion-based planning by incorporating a discriminative trajectory proposal module, which provides strong guidance for the generative refinement process. 
Furthermore, we design a fine-grained denoising decoder to enhance scene compliance, enabling more accurate trajectory prediction through enhanced alignment with the surrounding environment.
Experimental results demonstrate that DiffRefiner achieves state-of-the-art performance, attaining 87.4 \textit{EPDMS} on NAVSIM v2, and 87.1 \textit{DS} along with 71.4 \textit{SR} on Bench2Drive, thereby setting new records on both public benchmarks.
The effectiveness of each component is validated via ablation studies as well.
\end{abstract}

\begin{links}
    \link{Code}{https://github.com/nullmax-vision/DiffRefiner}
\end{links}


\section{Introduction}

\begin{figure}[t]
\centering
\includegraphics[width=1.0\linewidth]{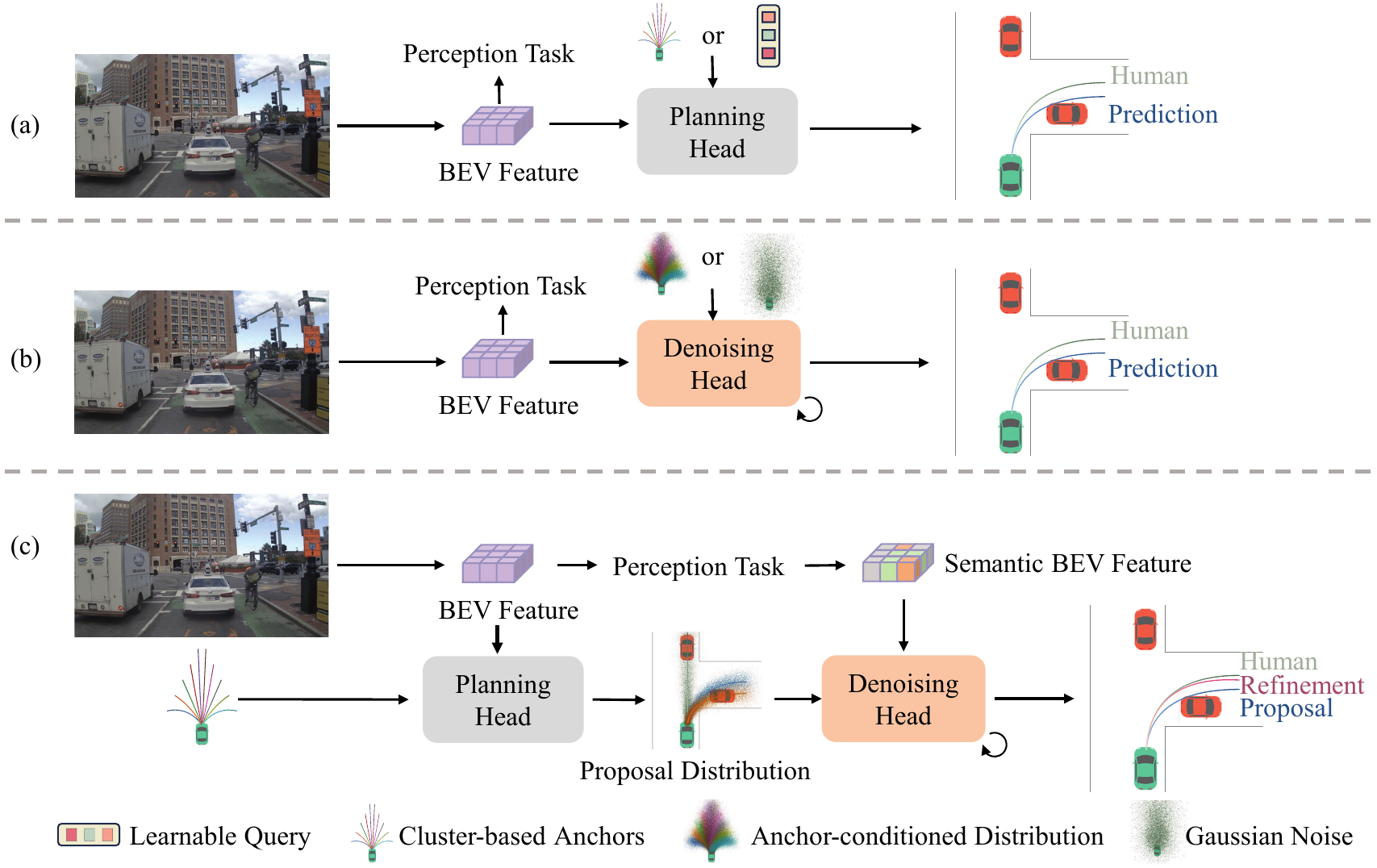}
\caption{Comparison of different paradigms for end-to-end planning: (a) single-stage discriminative approach, (b) single-stage generative diffusion method, and (c) our proposed coarse-to-fine framework integrates discriminative proposal construction with generative diffusion refinement.}
\label{fig:comparision}
\end{figure}

End-to-end autonomous driving (E2E-AD) has achieved significant progress in recent years, which directly maps raw sensor inputs to driving actions or trajectory planning~\cite{chitta2022transfuser, sun2024sparsedrive, chen2024end2end, weng2024paradrive, chitta2021neat, muhammad2020deep, hawke2020urban}. 
Compared to traditional approaches that rely on separated perception, prediction, and planning modules, end-to-end methods optimize the entire driving process in a unified manner, resulting in improved robustness and simpler deployment pipelines.

Prior approaches for ego-vehicle trajectory prediction typically employ single-pass regression on perception features or camera inputs~\cite{uniad,casas2021mp3,jiang2023vad,shao2023safety} (see Figure~\ref{fig:comparision} (a). While computationally efficient, these regression-based methods are fundamentally limited in their ability to handle the multimodal nature of trajectory prediction tasks.
The regression optimization process averages across multimodal behaviors, yielding suboptimal predictions particularly at complex intersections, and consequently exhibits poor generalization in real-world driving scenarios.
To address these challenges, recent work has investigated discretized solution spaces. For instance, several scoring-based approaches~\cite{chen2024vadv2, li2024hydra} employ offline-clustered trajectory anchors as discrete motion candidates, framing trajectory prediction as a classification problem. 
The distribution of future motions is then learned by evaluating each anchor either through its similarity to human demonstrations~\cite{chen2024vadv2} or via scores assigned from offline simulation-based assessments~\cite{yao2025drivesuprim, li2025hydraplus}. However, the computational complexity increases significantly with the size of the anchor set, limiting the feasibility of such methods for deployment in latency-sensitive autonomous driving systems~\cite{liao2025diffusiondrive}.

A growing research direction leverages diffusion models to address the multimodal challenges of driving behavior. Building on their remarkable success in image and video generation~\cite{yang2023diffusion,peebles2023scalable,kim2023neuralfield,yang2023law}, these methods show strong potential for trajectory prediction~\cite{motiondiff,graphdiff}. 
As shown in Figure~\ref{fig:comparision} (b), these models provide a continuous, generative framework for trajectory prediction. Through iterative denoising of Gaussian-distributed samples~\cite{chi2023diffusion}, they can generate diverse, physically plausible trajectories while naturally capturing the multimodality inherent in driving decisions.
DiffusionDrive~\cite{liao2025diffusiondrive} achieves this by generating diverse real-time trajectories through denoising of samples drawn from an anchor-parameterized Gaussian mixture model.
However, current diffusion-based approaches suffer from critical limitations in their initialization process. These methods rely on unstructured Gaussian noise ~\cite{zheng2025diffusion} or fixed trajectory-derived anchors ~\cite{bae2024singulartrajectory}, both of which lack scene adaptability. When the initial samples deviate from feasible motion distributions, this necessitates excessive denoising iterations, which in turn leads to increased computational latency.

To address these challenges, we propose DiffRefiner, a novel two-stage trajectory prediction framework that adopts a coarse-to-fine architecture.
A transformer-based Proposal Decoder first generates coarse trajectory predictions by regressing from a bank of predefined anchors, producing structured priors that serve as high-level guidance for subsequent refinement. The second stage employs a conditional Diffusion Refiner that iteratively refines these initial predictions via a generative diffusion process, thereby capturing trajectory details. 
This hybrid approach significantly improves diffusion-based planning performance by incorporating discriminative trajectory proposals that provide strong initialization for the subsequent generative refinement process.
To enhance scene compliance, we propose a fine-grained denoising decoder that employs constrained diffusion to achieve precise alignment between predicted trajectories and the surrounding environment.Specifically, we propose a Fine-Grained Semantic Interaction Module that systematically integrates environmental constraints into the trajectory refinement process. The module operates in three stages: first, cross-attention layers establish dense correspondences between trajectory features and BEV semantic regions (e.g., drivable areas and obstacles) to encode holistic global context and scene-level dependencies; second, deformable attention selectively aligns trajectory endpoints with critical region semantics to extract fine-grained local structures and interaction cues; and third, an adaptive gating network dynamically fuses the global scene representation with localized semantic information, enabling the model to balance coarse contextual understanding and precise spatial alignment. This hierarchical design facilitates accurate, context-aware trajectory optimization within the diffusion-based refinement framework.

In the experiments, we evaluate DiffRefiner on the open-loop real-world dataset NAVSIM and the closed-loop simulation benchmark Bench2Drive~\cite{jia2024bench2drive}. Experimental results demonstrate that DiffRefiner achieves state-of-the-art performance, attaining 87.4 \textit{EPDMS} on NAVSIM v2, and 87.1 \textit{DS} along with 71.4 \textit{SR} on Bench2Drive, thereby setting new records on both public benchmarks. 

The main contributions of the paper can be summarized as follows:

\begin{itemize}
    \item We propose a coarse-to-fine planning framework that first generates efficient anchor-based trajectory proposals as strong priors, then optimizes them through diffusion-based refinement.
    \item We present a fine-grained denoising decoder with a Scene-Aware Semantic Interaction Module that achieves precise trajectory refinement through optimized environment alignment during denoising. 
    \item Our DiffRefiner achieves state-of-the-art (SOTA) performance on the open-loop real-world benchmark NAVSIM v2 and the closed-loop simulation benchmark Bench2Drive.

\end{itemize}

\begin{figure*}[t]
\centering
\includegraphics[width=0.9\linewidth]{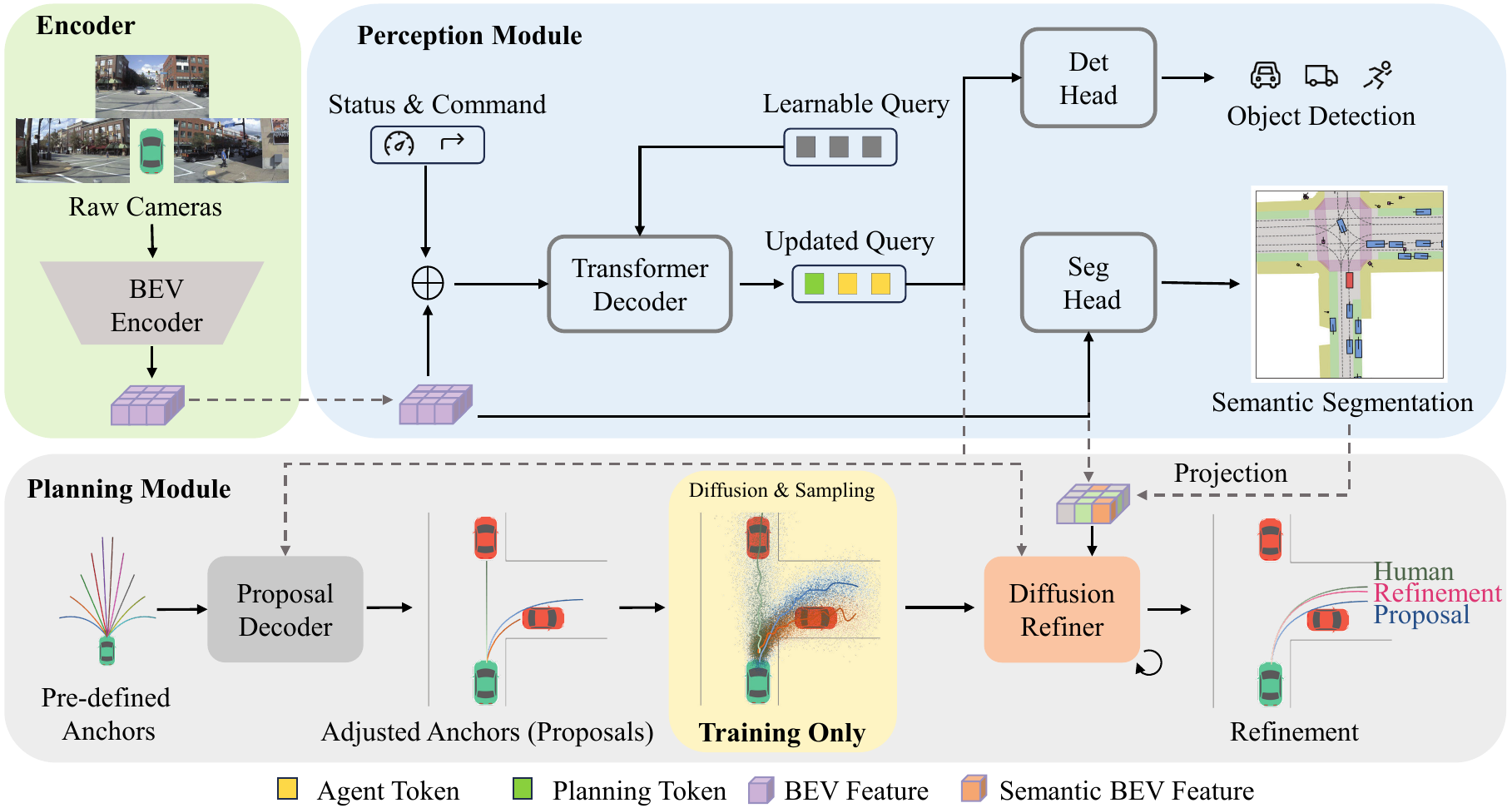}
\caption{Overview of the proposed DiffRefiner.
The DiffRefiner architecture comprises three primary components: a BEV encoder, an perception module, and a planning module, which sequentially perform scene representation learning, perception, and motion planning. The planning module is further decomposed into two submodules:
(a) a proposal decoder, which employs a discriminative approach to produce coarse proposals that capture the overall motion trend; and
(b) a diffusion refiner, which refines the proposals by leveraging a fine-grained denoising decoder conditioned on explicitly modeled scene semantics, thereby generating a final trajectory that better complies with environmental constraints.}
\label{fig:overview}
\end{figure*}

\section{Related Work}

\subsection{End-to-End Autonomous Driving}

Early end-to-end autonomous driving approaches~\cite{hawke2020urban,jiang2023vad,uniad} employ single-modal prediction, directly mapping sensor inputs to deterministic trajectories. Subsequent methods~\cite{chen2024vadv2,li2024hydra,li2025hydraplus,yao2025drivesuprim,sun2024sparsedrive} address the multimodal nature of human driving through discriminative frameworks, formulating trajectory prediction either as anchor-based classification (using predefined or clustered anchors) or as offset regression from these anchors.
Although these methods improve diversity over single-modal approaches, their performance remained limited by anchor coverage and the inability to model complex motion distributions.


Recently, diffusion-based generative methods have emerged as a promising alternative for trajectory prediction. Originally developed for image generation~\cite{rombach2022high,peebles2023scalable}, these approaches demonstrate strong capacity for modeling complex multi-modal motion distributions~\cite{kondo2024cgd}. By learning to reverse a Gaussian corruption process applied to ground truth trajectories~\cite{janner2022planning}, they generate diverse, physically plausible motions~\cite{zhu2023diffusion}. 
For improved efficiency, DiffusionDrive~\cite{liao2025diffusiondrive} employs a truncated diffusion process initialized from clustered trajectory anchors, while Diffusion Planner~\cite{zheng2025diffusion} combines transformer architectures with classifier guidance. Compared to deterministic regression, these diffusion-based approaches demonstrate superior ability to capture multimodality and maintain robustness in diverse driving scenarios.

\subsection{Coarse-To-Fine  Trajectory Planning}

The coarse-to-fine framework has become a predominant architecture for trajectory planning, effectively reducing search space while enabling progressive refinement~\cite{xing2025goalflow,jia2025drivetransformer,guo2025ipad}. These approaches first generate coarse trajectory candidates and subsequently refine them using specialized modules.
For example, some methods employ diverse planning strategies, including approaches that first detect key agents and construct scene representations before generating final outputs~\cite{su2024difsd}. Alternative frameworks utilize dual-decoder architectures to select coarse trajectory candidates for subsequent attention-based refinement~\cite{yao2025drivesuprim}. Additional methods enhance prediction accuracy by generating initial rough paths and refining them through iterative optimization~\cite{wang2024trajfine, xing2025goalflow}.

\subsection{Perception aware Trajectory Planning}

Early approaches to end-to-end autonomous driving~\cite{bojarski2016end} directly map sensor inputs to planning outputs without incorporating explicit perception modules. 
Subsequent methods~\cite{jiang2023vad,uniad,sun2024sparsedrive,weng2024paradrive}  substantially enhance planning performance through multi-task learning, where auxiliary perception tasks provide rich supervisory signals for the planning module. 
These methods typically process multi-modal sensor data through shared representations and leverage implicit feature interactions between perception and planning. 
Another line of work~\cite{ICLR2025_6aa49679,li2025end} investigates self-supervised perception through temporal modeling of sensory inputs. While these approaches achieve greater computational efficiency, they lack mechanisms for explicit semantic understanding of the driving environment.
However, a common limitation of these approaches is their inability to enable fine-grained interaction between perception and planning, which can result in unsafe behaviors such as collisions or violations of traffic constraints. 
In contrast, our framework establishes explicit semantic grounding for trajectory generation, where structured scene understanding provides fine-grained planning guidance to effectively mitigate these safety-critical failures.

\section{Method}
End-to-end autonomous driving takes raw sensor inputs and directly predicts the future trajectory of the ego vehicle. The predicted trajectory is denoted as $Y = \{{Y_i}\}_{i=1}^{T_f}$, where $T_f$ indicates the prediction horizon, and $Y_i$ represents the state of the ego vehicle at time step $i$, including its position and heading.

\subsection{Overall Framework}

The proposed DiffRefiner framework integrates three key components: a perception module, proposal decoder, and diffusion-based refiner, in a unified coarse-to-fine trajectory planning architecture.
As illustrated in Figure~\ref{fig:overview}, the framework consists of three major components: (1) a BEV-centric perception module that processes sensor inputs and is trained with auxiliary tasks 
to enhance scene understanding; (2) a coarse trajectory proposal decoder that employs a lightweight Transformer~\cite{vaswani2017attention} to adjust anchors and generate initial path predictions; and (3) a diffusion-based trajectory refiner that iteratively denoises and refines the proposals to produce optimized trajectories that better capture real-world driving complexity.

\subsection{Perception Module}

The perception module utilizes a BEV encoder~\cite{jaeger2023tf++} to generate bird's-eye-view (BEV) features \(F_{\text{bev}}\) from raw sensor inputs. The module processes these features through two complementary heads: a sparse agent head for detecting individual objects and a dense segmentation head for comprehensive scene understanding. This dual-head architecture enables simultaneous object-level agent understanding and pixel-wise environment segmentation.

The segmentation head transforms the BEV features through a semantic segmentation network:
\begin{equation}
\hat{\mathcal{S}} = \mathcal{F}{\text{seg}}(F_{\text{bev}}), 
\end{equation}
where \(\hat{\mathcal{S}}\) denotes the predicted semantic maps containing road elements, dynamic agents, and static obstacles. 

For the sparse agent computation, to incorporate the ego vehicle's state information, we encode its dynamic status (including velocity and acceleration) and navigation commands into a compact latent representation. This representation is combined with the scene context and processed by a transformer-based decoder that operates on a set of learnable queries \(Q\). The decoder produces updated queries that are divided into two distinct types: a \textit{Planning Token} \(T_p\) for trajectory generation and an \textit{Agent Token} \(T_a\) for sparse detection tasks. The detection head processes the agent token to predict surrounding objects:
\begin{equation}
\hat{\mathcal{D}} = \mathcal{F}_{\text{det}}\!\left(T_{a}\right),
\end{equation}
where \(\hat{\mathcal{D}}\) represents the detected agents' positions and categories in the environment.

\subsection{Proposal Decoder}

In the first stage, we employ a lightweight Transformer-based approach that predicts offsets to adjust the predefined anchors~\cite{li2025end}, yielding the adjusted anchors as trajectory proposals. The method takes a set of pre-defined trajectory anchors, typically obtained through offline clustering as discrete motion candidates, and predicts the trajectory output via a Transformer decoder.

We define the offline-clustered discrete trajectory vocabulary as $\mathcal{V}c$. Each anchor is position-encoded and projected by an MLP to form initial proposal queries, which are subsequently contextualized via cross-attention with the Planning Token $T_p$:
\begin{equation}
    Q_{\text{proposal}} = \text{CrossAttn}(Q=\text{MLP}(\text{pos}(\mathcal{V}c)), K=V=T_p)
\end{equation}
where $\text{pos}(\cdot)$ denotes sinusoidal positional encoding, $\text{MLP}(\cdot)$ projects each encoded anchor to a latent query space, $T_p$ provides the planning-aware context, and $Q_{\text{proposal}}$ represents the resulting context-enhanced trajectory queries.


\begin{figure}[t]
\centering
\includegraphics[width=0.95\linewidth]{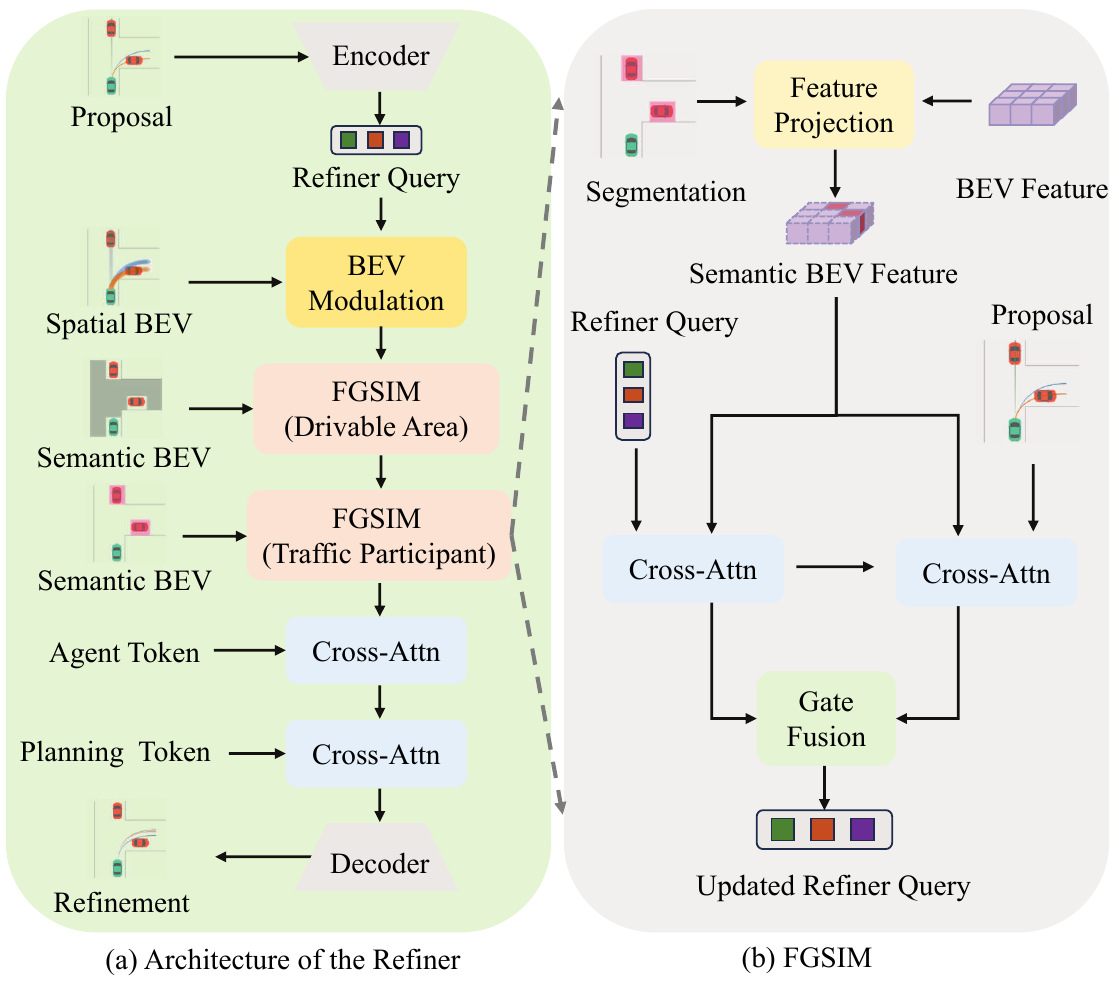}
\caption{Illustration of the detailed architecture of the refiner and the Fine-Grained Semantic Interaction Module (FGSIM).}
\label{fig:refiner}
\end{figure}

\subsection{Diffusion Refiner}
In the second stage, our diffusion-based refiner optimizes all trajectory proposals through conditional denoising, generating more realistic and context-aware predictions.
Specifically, as shown in Figure~\ref{fig:refiner}, we develop a fine-grained denoising decoder that explicitly enforces trajectory-environment alignment constraints during the iterative refinement process, ensuring enhanced compliance with scene semantics and dynamics.


\paragraph{Training Phase.}  
During training, we simulate the forward diffusion process~\cite{ho2020denoising}  by progressively adding Gaussian noise to \( Y_{\text{proposal}} \) over \( T \) steps. At a randomly sampled step \( t \), the noisy trajectory \( \tilde{Y} \) is computed as:
\begin{equation}
\tilde{Y} = \sqrt{\bar{\alpha}_t} Y_{\text{proposal}} + \sqrt{1 - \bar{\alpha}_t} \epsilon,\quad \epsilon \sim \mathcal{N}(0, \mathbf{I}),
\end{equation}
where \( \bar{\alpha}_t \) is the cumulative product of the noise schedule~\cite{song2021denoising} . The noisy sample \( \tilde{Y} \) is encoded into refinement queries through positional encoding and MLPs as a multi-modal ego query:
\begin{equation}
Q_{\text{refiner}} = \text{Enc}(\tilde{Y}) = \text{MLP}(\text{pos}(\tilde{Y}))
\end{equation}



\paragraph{Conditional Denoising with Scene-Aware Interaction.}
The scene-aware conditional denoising module enhances trajectory prediction through a hierarchical refinement process that integrates multi-level environmental context, as illustrated in Figure~\ref{fig:refiner}. 


The architecture begins with a spatial BEV modulation module~\cite{liao2025diffusiondrive} that extracts trajectory-conditioned spatial features from the BEV representation. These features initialize our Fine-Grained Semantic Interaction Module (FGSIM), which operates through two sequential refinement stages.

The first stage performs road-aware refinement by combining refiner queries with BEV features and drivable area segmentation, thereby constraining predictions to physically navigable road regions. Building upon this, the second stage conducts interaction-aware refinement by incorporating dynamic agent features, enabling explicit modeling of traffic participant interactions and proactive collision avoidance.

Subsequent cross-attention layers further refine predictions by capturing inter-agent relationships and ego-vehicle motion constraints. Finally, parallel MLP heads predict the refined trajectories and their confidence score, ensuring physical feasibility and contextual consistency.

\paragraph{Fine-Grained Semantic Interaction Module}

As map-based and interaction-based reasoning represent two fundamental aspects of autonomous planning, we introduce a semantic-aware interaction module that explicitly accounts for both. The module sequentially enhances the planner by aligning it with map semantics and dynamic agent interactions, while hierarchically integrating global scene context and local structural cues into trajectory decoding to improve scene understanding.
To enable such targeted interaction, the module first extracts semantically critical regions from the perception output, providing region-level guidance for subsequent map- and agent-based reasoning.

To identify critical regions that are highly relevant to downstream planning, we leverage the semantic segmentation output $\hat{\mathcal{S}}$ to extract semantically meaningful areas of interest:
\begin{equation}
\mathcal{R} = \big\{ \{ r_{ij} \}_{i=1}^{I} \big\}_{j=1}^{J},
\end{equation}
where \( r_{ij} \) denotes the \( i \)-th region of the \( j \)-th semantic category (e.g., lane boundaries, crosswalks). 
Category-specific semantic masks are applied to filter the segmentation map, and the resulting regions are projected into the BEV feature space to enable region-aware feature extraction:
\begin{equation}
F_R = \big\{ \operatorname{proj}(F_{\text{bev}}, \mathcal{R}_j) \big\}_{j=1}^{J}.
\end{equation}

The module then processes each semantic feature \( F_{R_i} \) through complementary attention mechanisms designed to jointly capture global scene context and local geometric details. Here, \( Q_{\text{refiner}} \) denotes the updated refiner queries passed from the previous interaction stage. This is achieved via a two-step attention process: first, a global cross-attention operation integrates scene-wide contextual information, followed by a local deformable attention mechanism that adaptively focuses on regions spatially relevant to the trajectory:
\begin{align}
Q_r^{(c)} &= \operatorname{CrossAttn}(Q=Q_{\text{refiner}}, K=V=F_{R_i}), \\
Q_r^{(d)} &= \operatorname{DeformAttn}(Q=Q_r^{(c)}, K=V=F_{R_i}; \tilde{Y}_{T}),
\end{align}
where \( \tilde{Y}_T \) provides trajectory-adaptive spatial reference.

A gated fusion mechanism dynamically balances these representations:
\begin{align}
\text{Gate} &= \sigma\big(W_{\text{gate}}(Q_r^{(c)}, Q_r^{(d)})\big), \\
Q_r &= Q_r^{(c)} \cdot \text{Gate} + Q_r^{(d)} \cdot (1-\text{Gate}),
\end{align}
where $\sigma(\cdot)$ denotes the sigmoid activation function that maps the input to the range $(0,1)$, 
$W_{\text{gate}}$ is a learnable linear projection used to compute the gating coefficient, 
and $Q_r$ denotes the updated refiner query after fusing global and local interactions.

\begin{table*}[t]
    \centering
    \setlength{\tabcolsep}{2pt}
    \begin{tabular}{c|c|c|c|cccc|cccccc}
        \toprule
        Method & Backbone & Modality & EPDMS$\uparrow$ & NC$\uparrow$ & DAC$\uparrow$ & DDC$\uparrow$ & TL$\uparrow$ & EP$\uparrow$ & TTC$\uparrow$ & LK$\uparrow$ & HC$\uparrow$ & EC$\uparrow$ \\
        \midrule
        Human Agent & - & - & 90.3 & 100 & 100 & 99.8 & 100 & 87.4 & 100 & 100 & 98.1 & 90.1 \\
        \midrule
        Transfuser~\cite{chitta2022transfuser} & ResNet34 & C+L & 76.7 & 96.9 & 89.9 & 97.8 & 99.7 & 87.1 & 95.4 & 92.7 & 98.3 & \textbf{87.2} \\
        DiffusionDrive\textsuperscript{*}~\cite{liao2025diffusiondrive} & ResNet34 & C+L & 84.0 & 98.2 & 96.2 & 98.6 & - & 87.6 & 97.3 & 97.4 & \textbf{98.4} & - \\
        GaussianFusion\textsuperscript{*}~\cite{liu2025gaussianfusion} & ResNet34 & C+L & 85.0 & 98.3 & 97.3 & 99.0 & - & 87.5 & 97.4 & 97.4 & 98.3 & - \\
        HydraMDP++~\cite{li2025hydraplus} & ResNet34 & C & 81.4 & 97.2 & \textbf{97.5} & 99.4 & 99.6 & 83.1 & 96.5 & 94.4 & 98.2 & 70.9 \\
        DriveSuprim~\cite{yao2025drivesuprim} & ResNet34 & C & 83.1 & 97.5 & 96.5 & 99.4 & 99.6 & \textbf{88.4} & 96.6 & 95.5 & 98.3 & 77.0 \\
        \textbf{DiffRefiner (Ours)} & ResNet34 & C & \textbf{86.2} & \textbf{98.5} & 97.4 & \textbf{99.6} & \textbf{99.8} & 87.6 & \textbf{97.7} & \textbf{97.7} & 98.3 & 86.2 \\
        \midrule
        HydraMDP++~\cite{li2025hydraplus} & V2-99 & C & 85.1 & 98.4 & 98.0 & 99.4 & 99.8 & 87.5 & 97.7 & 95.3 & \textbf{98.3} & 77.4 \\
        DriveSuprim~\cite{yao2025drivesuprim} & V2-99 & C & 86.0 & 97.8 & 97.9 & 99.5 & 99.9 & \textbf{90.6} & 97.1 & 96.6 & \textbf{98.3} & 77.9 \\
        \textbf{DiffRefiner (Ours)}  & V2-99 & C & \textbf{87.4} & \textbf{98.6} & \textbf{98.5} & \textbf{99.7} & \textbf{99.9} & 87.6 & \textbf{98.1} & \textbf{98.1} & \textbf{98.3} & \textbf{87.9} \\
        \bottomrule
    \end{tabular}
    \caption{
    Evaluation results on the NAVSIM v2 benchmark. 
    We report the overall score (EPDMS) and a set of detailed driving metrics. Results are grouped by backbone types (ResNet34 and V2-99). 
    Our proposed DiffRefiner consistently outperforms prior methods across most metrics and achieves the best overall performance. 
    \textsuperscript{*} indicates results reported from GaussianFusion~\cite{liu2025gaussianfusion}, while results of other baselines are from DriveSuprim~\cite{yao2025drivesuprim}.
    }
    \label{tab:results-navsimv2}
\end{table*}

\begin{table*} 
\centering
\setlength{\tabcolsep}{3pt}  
\begin{tabular}{c|c|cc|ccccc|c}
    \toprule
    \multirow{2}{*}{Method} & \multirow{2}{*}{Paradigm} & \multicolumn{2}{c|}{Overall$\uparrow$} & \multicolumn{6}{c}{Multi-Ability$\uparrow$} \\
    \cmidrule(lr){3-10}
    & & DS & SR(\%) & Merge & Overtake & EBrake & GiveWay & Tsign & Mean \\
    \midrule
    PDM-Lite~\cite{sima2024drivelm} & Rule based & 97.0 & 92.3 & 88.8 & 93.3 & 98.3 & 90.0 & 93.7 & 92.8 \\
    \midrule
    VAD~\cite{jiang2023vad} & Discriminative & 42.4 & 15.0 & 8.1 & 24.4 & 18.6 & 20.0 & 19.2 & 18.1 \\
    UniAD~\cite{uniad} & Discriminative & 45.8 & 16.4 & 14.1 & 17.8 & 21.7 & 10.0 & 14.2 & 15.6 \\
    ThinkTwice~\cite{jia2023think} & Discriminative & 62.4 & 33.2 & 27.4 & 18.4 & 35.8 & 50.0 & 54.4 & 37.2 \\
    DriveTransformer~\cite{jia2025drivetransformer} & Discriminative & 63.5 & 35.0 & 17.6 & 35.0 & 48.4 & 40.0 & 52.1 & 38.6 \\
    TF++~\cite{zimmerlin2024tf++dataset} & Discriminative & 84.2 & 67.3 & 58.8 & 57.8 & 83.3 & 40.0 & 82.1 & 64.4 \\
    HiPAD~\cite{tang2025hip} & Discriminative & 86.8 & 69.1 & 50.0 & \textbf{84.4} & 83.3 & 40.0 & 72.1 & 66.0 \\
    \midrule
    Orion~\cite{fu2025orion} & Generative & 77.7 & 54.6 & 25.0 & 71.1 & 78.3 & 30.0 & 69.2 & 54.7 \\
    GaussianFusion~\cite{liu2025gaussianfusion} & Generative & 79.4 & 59.5 & 40.0 & 66.7 & 66.7 & 50.0 & 63.7 & 57.4 \\
    \midrule
    \textbf{DiffRefiner (Ours)}  & Hybrid & \textbf{87.1} & \textbf{71.4} & \textbf{63.8} & 60.0 & \textbf{85.0} & \textbf{50.0} & \textbf{86.3} & \textbf{69.0} \\
    \bottomrule
\end{tabular}
\caption{
Performance comparison on the Bench2Drive benchmark. 
Our proposed DiffRefiner achieves the best overall performance among existing approaches, demonstrating significant improvements in DS, SR, and most multi-ability tasks.
}
\label{tab:resultscarla}
\end{table*}

\subsection{Training Loss}
\noindent
Following prior end-to-end approaches~\cite{uniad,jiang2023vad}, we employ a two-stage training scheme to enhance optimization stability.

In the first stage, the perception network is optimized using a Transfuser-style~\cite{chitta2022transfuser} perception loss, denoted as \(\mathcal{L}_{\mathrm{perception}}\).  

In the second stage, perception and planning are jointly optimized end-to-end. A winner-takes-all strategy selects the trajectory closest to the ground truth, and ego prediction loss is computed as:
\begin{equation}
\mathcal{L}_{\mathrm{planning}} = w_{\mathrm{reg}} \mathcal{L}_{\mathrm{reg}} + w_{\mathrm{cls}} \mathcal{L}_{\mathrm{cls}},
\end{equation}
where \(\mathcal{L}_{\mathrm{reg}}\) is the L1 regression loss and \(\mathcal{L}_{\mathrm{cls}}\) is the binary cross-entropy classification loss.
The final objective combines all components:
\begin{equation}
\mathcal{L}_{\mathrm{total}} = \mathcal{L}_{\mathrm{proposal}} + \mathcal{L}_{\mathrm{refinement}} + \mathcal{L}_{\mathrm{perception}},
\end{equation}
where \(\mathcal{L}_{\mathrm{proposal}}\) and \(\mathcal{L}_{\mathrm{refinement}}\) are the planning losses from the proposal and refinement modules.

\section{Experiments}
\subsection{Experimental Setup}
We use two widely recognized benchmarks: NAVSIM v2~\cite{cao2025navsimv2} for open-loop evaluation and Bench2Drive~\cite{jia2024bench2drive} for closed-loop testing.
\paragraph{NAVSIM.}
NAVSIM~\cite{dauner2024navsim}, based on the Openscene~\cite{peng2023openscene} dataset, is a real-world, planning-focused benchmark for evaluating autonomous driving models in open-loop settings. 
We evaluate on the \textit{Navtest} split, comprising 12,146 frames covering diverse scenarios such as intersections, dynamic agents, and varied traffic conditions.
To measure planning performance, we use the Extended Predictive Driver Model Score (EPDMS) introduced in NAVSIM v2~\cite{cao2025navsimv2}.
More details on the metrics are provided in Appendix B.

\paragraph{Bench2Drive.}
Bench2Drive~\cite{jia2024bench2drive} is a closed-loop evaluation benchmark based on CARLA~\cite{dosovitskiy2017carla}, designed to assess end-to-end autonomous driving systems in interactive urban scenarios. We evaluate our model on 220 routes spanning 44 diverse, interactive scenarios. Official metrics include Driving Score (DS), Success Rate (SR), and Multi-Ability Score, which collectively measure navigation performance, safety, and rule adherence.
For detailed metric definitions, see Appendix B.

\paragraph{Implementation Details.}

For the NAVSIM benchmark, we use the standard \textit{navtrain} split for training. Consistent with the NAVSIM v2 Challenge specifications, our model processes synchronized multi-view inputs from the front, left-front, and right-front cameras. We train with a batch size of 384 and a learning rate of 4e-4 for 100 epochs, applying identical training schedules for both the perception pretraining and end-to-end fine-tuning phases.
For Bench2Drive evaluation, we follow the dataset configuration and preprocessing pipeline established in TF++~\cite{zimmerlin2024tf++dataset}.
Both benchmarks employ 20 clustered trajectory anchors for proposal initialization follow DiffusionDrive~\cite{liao2025diffusiondrive}, and all proposals are then passed to the refinement module. All experiments are conducted on a cluster of 8 NVIDIA RTX 4090 GPUs. 
Additional implementation details are provided in Appendix C.


\subsection{Comparison with State-of-the-Art Methods}
\paragraph{Results on NAVSIM.}

We perform comprehensive open-loop evaluations on the NAVSIM v2 benchmark. As demonstrated in Table~\ref{tab:results-navsimv2}, our framework establishes new state-of-the-art performance, surpassing the previous best method (DriveSuprem~\cite{yao2025drivesuprim}) by significant margins of 3.7\% (ResNet34 backbone) and 1.6\% (V2-99 backbone). The results show particular improvements in safety critical metrics,
validating the efficacy of our architectural design.



\paragraph{Results on Bench2Drive.}
We further evaluate our approach in a closed-loop setting on the Bench2Drive benchmark.
As shown in Table~\ref{tab:resultscarla}, our method outperforms all existing learning-based baselines, achieving state-of-the-art results. Without model ensembling, it improves DS by 0.3 and SR by 2.3 over the previous best, HiPAD~\cite{tang2025hip}. Unlike prior discriminative or fully generative methods, our framework employs a hybrid generative paradigm that integrates coarse anchor-based proposals with diffusion-based refinement. This design yields consistent gains across most multi-ability metrics, demonstrating robustness and effectiveness in diverse interactive driving scenarios.

\begin{table}
    \centering
    \fontsize{9pt}{9pt}\selectfont
    \setlength{\tabcolsep}{4pt}  
    \begin{tabular}{c|cc|cc| c|c|c}
        \toprule
        ID & Ref & Pro & RT & Src & Param & Latency (ms) & EPDMS$\uparrow$ \\
        \midrule
        1 &   & \checkmark   &  -   & Pro & 57.2M & 12  & 85.0\\ 
        2 & \checkmark &  & Gen     & Ref & 73.7M  & 40  & 86.0 \\
        \midrule
        3 & \checkmark & \checkmark  & Dis   & Pro & 74.8M  & 12  & 85.5 \\   
        4 & \checkmark & \checkmark  & Dis   & Ref & 74.8M  & 27  & 78.3 \\   
        5 & \checkmark & \checkmark  & Gen   & Pro & 74.8M  & 12  & 85.8 \\   
        6 & \checkmark & \checkmark  & Gen   & Ref & 74.8M  & 27  & \textbf{86.2} \\   
        \bottomrule
    \end{tabular}
    \caption{Ablation study of the proposed planning framework. 
        Ref: Refinement; Pro: Proposal; 
        RT: Refiner type; Src: Input source of downstream control; 
        Gen: Generative; Dis: Discriminative; 
        Param: Total number of model parameters; 
        Latency: End-to-end planning latency.}
    \label{tab:ablation of framework}
\end{table}

\begin{table}[t] 
    \centering
    \fontsize{9pt}{9pt}\selectfont
    \begin{tabular}{c|ccc|cc|ccc}
        \toprule
        ID & P & A & M & DA & TP & EPDMS$\uparrow$ & NC$\uparrow$ & DAC$\uparrow$ \\
        \midrule
        1 & \checkmark &         &        &        &        &  82.4 & 97.9  & 93.8  \\
        2 & \checkmark & \checkmark &    &        &        &  82.9 & 98.0  & 94.5  \\
        3 & \checkmark &  & \checkmark   &        &        &  83.3 & 98.0  & 94.9  \\
        4 & \checkmark & \checkmark & \checkmark &        &        &  83.5 & 98.1  & 95.1  \\
        5 & \checkmark & \checkmark & \checkmark & \checkmark &        &  84.3 & 98.0  & 95.7  \\
        6 & \checkmark & \checkmark & \checkmark & \checkmark & \checkmark &  \textbf{85.0} & \textbf{98.4}  & \textbf{96.3}  \\
        \bottomrule
    \end{tabular}
    \caption{Ablation study of refiner components. 
        P: Planning token; \textbf{A}: Agent token; 
        M: BEV modulation; 
        DA: Drivable area in FGSIM; 
        TP: Traffic participant in FGSIM.}

    \label{tab:ablation of Components}
\end{table}

\begin{table}
    \centering
    \fontsize{9pt}{9pt}\selectfont
    \setlength{\tabcolsep}{4pt}  
    \begin{tabular}{c|cc|c|ccc}
        \toprule
        ID & Global & Local & Fusion & EPDMS$\uparrow$ & NC$\uparrow$ & DAC$\uparrow$ \\
        \midrule
        1 & \checkmark &         & -    & 85.9 & 98.4 & 97.2 \\
        2 &            & \checkmark & -    & 85.9 & 98.4 & 97.2 \\
        3 & \checkmark & \checkmark & Addition  & 85.9 & 98.4 & 97.2 \\
        4 & \checkmark & \checkmark & Gating & \textbf{86.2} & \textbf{98.5} & \textbf{97.4} \\
        \bottomrule
    \end{tabular}
    \caption{Ablation study of FGSIM components. 
        ``Global'' and ``Local'' refer to global cross-attention and local deformable attention, respectively. 
        ``Fusion'' compares additive and gating-based fusion strategies.}
    \label{tab:ablation of FGSIM}
\end{table}

\subsection{Ablation Study}

\paragraph{Ablation on Planning Framework.}
The impact of each stage in our planning framework is evaluated in Table~\ref{tab:ablation of framework}. A comparison between rows 1 and 6 indicates that incorporating the refiner improves EPDMS by 1.2, confirming that it substantially enhances the quality of trajectory proposals. Rows 2 and 6 further show that higher-quality proposals raise the refinement upper bound, 
where row~2 applies two denoising iterations while row~6 uses a single iteration. 
Meanwhile, rows 1 and 5 indicate that proper refiner supervision benefits proposal learning.
Finally, the comparison of rows 4 and 6 highlights the advantage of our generative refiner over discriminative alternatives in performing fine-grained trajectory adjustments.

\paragraph{Ablation on Refiner Components.}

As shown in Table~\ref{tab:ablation of Components}, we perform a systematic ablation to assess the contribution of each module in the refiner decoder. Results show consistent gains from all components, validating the overall design.
Comparing rows 4–6, the semantic interaction mechanism progressively improves scene understanding and mitigates collision-related errors by exploiting fine-grained semantic cues.


\paragraph{Analysis of FGSIM Components.}
As illustrated in Table~\ref{tab:ablation of FGSIM}, both global context and local target cues independently lead to performance improvements, demonstrating their complementary roles. However, a naive additive fusion of the two results in performance degradation due to conflicting information. In contrast, our gating mechanism adaptively balances their contributions, achieving the best performance and confirming the benefit of adaptive feature integration.

\begin{figure}[t]
\centering
\includegraphics[width=0.9\linewidth]{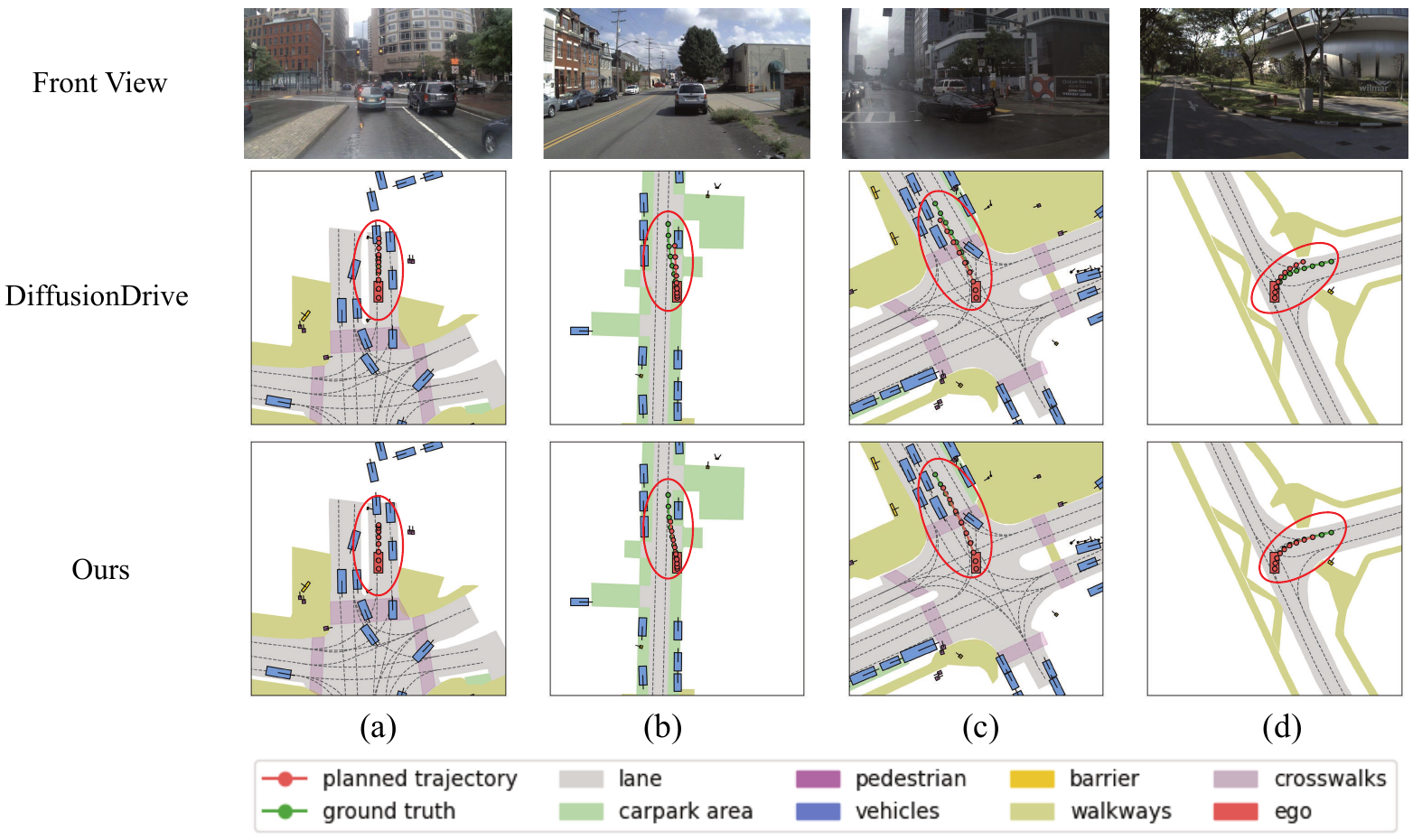}
\caption{Visualization of representative examples of DiffusionDrive~\cite{liao2025diffusiondrive} and our method. (a) and (b) illustrate cases in which our method achieves better collision avoidance compared with DiffusionDrive, whereas (c) and (d) demonstrate cases where our method exhibits improved compliance with lane constraints.}
\label{fig:visable}
\end{figure}

\paragraph{Ablation on Denoising Steps.}
The results in Table~\ref{tab:ablation of step} demonstrate that near-optimal performance can be achieved with just a single denoising step. 
This finding underscores the effectiveness of high-quality proposals as strong priors, enabling efficient diffusion-based refinement and highlighting the suitability of our framework for real-time end-to-end autonomous driving.

\begin{table}
    \centering
    \fontsize{9pt}{9pt}\selectfont
    \setlength{\tabcolsep}{4pt}  
    \begin{tabular}{c|c|cccccc}
        \toprule
        Steps & EPDMS$\uparrow$ & NC$\uparrow$ & DAC$\uparrow$ & DDC$\uparrow$ & EP$\uparrow$ & LK$\uparrow$ \\
        \midrule
        1 & 86.20 & 98.47 & 97.36 & \textbf{99.64} & \textbf{87.59} & \textbf{97.79} \\
        2 & \textbf{86.22} & 98.47 & \textbf{97.37} & 99.63  & 87.58 & 97.74 \\
        5 & 86.17 & \textbf{98.48} & 97.34 & 99.63  & 87.58  & 97.78 \\
        \bottomrule
    \end{tabular}
    \caption{
    Ablation study on the number of denoising steps.  
    }
    \label{tab:ablation of step}
\end{table}

\subsection{Qualitative Analysis}
As illustrated in Figure~\ref{fig:visable}, our method surpasses DiffusionDrive~\cite{liao2025diffusiondrive} in complex interactive scenarios by better attending to fine-grained scene details. It reduces collisions with surrounding agents and adheres more strictly to map constraints, resulting in higher-quality trajectories. 
Additional qualitative results, including closed-loop simulation cases, are provided in Appendix E.

\section{Conclusions}
In this work, we present DiffRefiner, a novel two-stage trajectory prediction framework for end-to-end autonomous driving planning.  
DiffRefiner incorporates a transformer-based proposal decoder to generate coarse trajectories, which provide strong guidance for subsequent generative refinement. The diffusion refiner further improves these proposals via iterative denoising with a fine-grained decoder, thereby enhancing scene compliance and producing more accurate and realistic trajectories. Extensive experiments on NAVSIM v2 and Bench2Drive demonstrate that DiffRefiner achieves state-of-the-art performance.

\bibliography{aaai2026}

@article{liu2025gaussianfusion,
  title={GaussianFusion: Gaussian-Based Multi-Sensor Fusion for End-to-End Autonomous Driving},
  author={Liu, Shuai and Liang, Quanmin and Li, Zefeng and Li, Boyang and Huang, Kai},
  journal={arXiv preprint arXiv:2506.00034},
  year={2025}
}

@inproceedings{sima2024drivelm,
  title={Drivelm: Driving with graph visual question answering},
  author={Sima, Chonghao and Renz, Katrin and Chitta, Kashyap and Chen, Li and Zhang, Hanxue and Xie, Chengen and Bei{\ss}wenger, Jens and Luo, Ping and Geiger, Andreas and Li, Hongyang},
  booktitle={European conference on computer vision},
  pages={256--274},
  year={2024},
  organization={Springer}
}

@inproceedings{jia2023think,
  title={Think twice before driving: Towards scalable decoders for end-to-end autonomous driving},
  author={Jia, Xiaosong and Wu, Penghao and Chen, Li and Xie, Jiangwei and He, Conghui and Yan, Junchi and Li, Hongyang},
  booktitle={Proceedings of the IEEE/CVF Conference on Computer Vision and Pattern Recognition},
  pages={21983--21994},
  year={2023}
}

@inproceedings{
jia2025drivetransformer,
title={DriveTransformer: Unified Transformer for Scalable End-to-End Autonomous Driving},
author={Xiaosong Jia and Junqi You and Zhiyuan Zhang and Junchi Yan},
booktitle={The Thirteenth International Conference on Learning Representations},
year={2025},
url={https://openreview.net/forum?id=M42KR4W9P5}
}

@article{fu2025orion,
  title={Orion: A holistic end-to-end autonomous driving framework by vision-language instructed action generation},
  author={Fu, Haoyu and Zhang, Diankun and Zhao, Zongchuang and Cui, Jianfeng and Liang, Dingkang and Zhang, Chong and Zhang, Dingyuan and Xie, Hongwei and Wang, Bing and Bai, Xiang},
  journal={arXiv preprint arXiv:2503.19755},
  year={2025}
}

@article{tang2025hip,
  title={Hip-ad: Hierarchical and multi-granularity planning with deformable attention for autonomous driving in a single decoder},
  author={Tang, Yingqi and Xu, Zhuoran and Meng, Zhaotie and Cheng, Erkang},
  journal={arXiv preprint arXiv:2503.08612},
  year={2025}
}

@inproceedings{
song2021denoising,
title={Denoising Diffusion Implicit Models},
author={Jiaming Song and Chenlin Meng and Stefano Ermon},
booktitle={International Conference on Learning Representations},
year={2021},
url={https://openreview.net/forum?id=St1giarCHLP}
}

@misc{motiondiff,
      title={MotionDiffuser: Controllable Multi-Agent Motion Prediction using Diffusion},
      author={Chiyu Max Jiang and Andre Cornman and Cheolho Park and Ben Sapp and Yin Zhou and Dragomir Anguelov},
      year={2023},
      eprint={2306.03083},
      archivePrefix={arXiv},
      primaryClass={cs.RO}
}

@inproceedings{uniad,
  title={Planning-oriented autonomous driving},
  author={Hu, Yihan and Yang, Jiazhi and Chen, Li and Li, Keyu and Sima, Chonghao and Zhu, Xizhou and Chai, Siqi and Du, Senyao and Lin, Tianwei and Wang, Wenhai and others},
  booktitle={Proceedings of the IEEE/CVF conference on computer vision and pattern recognition},
  pages={17853--17862},
  year={2023}
}

@inproceedings{casas2021mp3,
  title={Mp3: A unified model to map, perceive, predict and plan},
  author={Casas, Sergio and Sadat, Abbas and Urtasun, Raquel},
  booktitle={Proceedings of the IEEE/CVF Conference on Computer Vision and Pattern Recognition},
  pages={14403--14412},
  year={2021}
}

@inproceedings{jiang2023vad,
  title={Vad: Vectorized scene representation for efficient autonomous driving},
  author={Jiang, Bo and Chen, Shaoyu and Xu, Qing and Liao, Bencheng and Chen, Jiajie and Zhou, Helong and Zhang, Qian and Liu, Wenyu and Huang, Chang and Wang, Xinggang},
  booktitle={Proceedings of the IEEE/CVF International Conference on Computer Vision},
  pages={8340--8350},
  year={2023}
}

@article{sun2024sparsedrive,
  title={Sparsedrive: End-to-end autonomous driving via sparse scene representation},
  author={Sun, Wenchao and Lin, Xuewu and Shi, Yining and Zhang, Chuang and Wu, Haoran and Zheng, Sifa},
  journal={arXiv preprint arXiv:2405.19620},
  year={2024}
}

@article{yao2025drivesuprim,
  title={DriveSuprim: Towards Precise Trajectory Selection for End-to-End Planning},
  author={Yao, Wenhao and Li, Zhenxin and Lan, Shiyi and Wang, Zi and Sun, Xinglong and Alvarez, Jose M and Wu, Zuxuan},
  journal={arXiv preprint arXiv:2506.06659},
  year={2025}
}

@article{chitta2022transfuser,
  title={Transfuser: tion with transformer-based sensor fusion for autonomous driving},
  author={Chitta, Kashyap and Prakash, Aditya and Jaeger, Bernhard and Yu, Zehao and Renz, Katrin and Geiger, Andreas},
  journal={IEEE transactions on pattern analysis and machine intelligence},
  volume={45},
  number={11},
  pages={12878--12895},
  year={2022},
  publisher={IEEE}
}

@article{chen2024end2end,
  title={End-to-end autonomous driving: Challenges and frontiers},
  author={Chen, Li and Wu, Penghao and Chitta, Kashyap and Jaeger, Bernhard and Geiger, Andreas and Li, Hongyang},
  journal={IEEE Transactions on Pattern Analysis and Machine Intelligence},
  year={2024},
  publisher={IEEE}
}

@inproceedings{shao2023safety,
  title={Safety-enhanced autonomous driving using interpretable sensor fusion transformer},
  author={Shao, Hao and Wang, Letian and Chen, Ruobing and Li, Hongsheng and Liu, Yu},
  booktitle={Conference on Robot Learning},
  pages={726--737},
  year={2023},
  organization={PMLR}
}

@inproceedings{jaeger2023tf++,
  title={Hidden biases of end-to-end driving models},
  author={Jaeger, Bernhard and Chitta, Kashyap and Geiger, Andreas},
  booktitle={Proceedings of the IEEE/CVF International Conference on Computer Vision},
  pages={8240--8249},
  year={2023}
}

@article{zimmerlin2024tf++dataset,
  title={Hidden biases of end-to-end driving datasets},
  author={Zimmerlin, Julian and Bei{\ss}wenger, Jens and Jaeger, Bernhard and Geiger, Andreas and Chitta, Kashyap},
  journal={arXiv preprint arXiv:2412.09602},
  year={2024}
}

@article{muhammad2020deep,
  title={Deep learning for safe autonomous driving: Current challenges and future directions},
  author={Muhammad, Khan and Ullah, Amin and Lloret, Jaime and Del Ser, Javier and De Albuquerque, Victor Hugo C},
  journal={IEEE Transactions on Intelligent Transportation Systems},
  volume={22},
  number={7},
  pages={4316--4336},
  year={2020},
  publisher={IEEE}
}

@article{chen2024vadv2,
  title={Vadv2: End-to-end vectorized autonomous driving via probabilistic planning},
  author={Chen, Shaoyu and Jiang, Bo and Gao, Hao and Liao, Bencheng and Xu, Qing and Zhang, Qian and Huang, Chang and Liu, Wenyu and Wang, Xinggang},
  journal={arXiv preprint arXiv:2402.13243},
  year={2024}
}

@article{li2024hydra,
  title={Hydra-mdp: End-to-end multimodal planning with multi-target hydra-distillation},
  author={Li, Zhenxin and Li, Kailin and Wang, Shihao and Lan, Shiyi and Yu, Zhiding and Ji, Yishen and Li, Zhiqi and Zhu, Ziyue and Kautz, Jan and Wu, Zuxuan and others},
  journal={arXiv preprint arXiv:2406.06978},
  year={2024}
}

@inproceedings{hawke2020urban,
  title={Urban driving with conditional imitation learning},
  author={Hawke, Jeffrey and Shen, Richard and Gurau, Corina and Sharma, Siddharth and Reda, Daniele and Nikolov, Nikolay and Mazur, Przemys{\l}aw and Micklethwaite, Sean and Griffiths, Nicolas and Shah, Amar and others},
  booktitle={2020 IEEE International Conference on Robotics and Automation (ICRA)},
  pages={251--257},
  year={2020},
  organization={IEEE}
}

@article{chi2023diffusion,
  title={Diffusion policy: Visuomotor policy learning via action diffusion},
  author={Chi, Cheng and Xu, Zhenjia and Feng, Siyuan and Cousineau, Eric and Du, Yilun and Burchfiel, Benjamin and Tedrake, Russ and Song, Shuran},
  journal={The International Journal of Robotics Research},
  pages={02783649241273668},
  year={2023},
  publisher={SAGE Publications Sage UK: London, England}
}

@inproceedings{weng2024paradrive,
  title={Para-drive: Parallelized architecture for real-time autonomous driving},
  author={Weng, Xinshuo and Ivanovic, Boris and Wang, Yan and Wang, Yue and Pavone, Marco},
  booktitle={Proceedings of the IEEE/CVF Conference on Computer Vision and Pattern Recognition},
  pages={15449--15458},
  year={2024}
}

@article{yang2023diffusion,
  title={Diffusion probabilistic modeling for video generation},
  author={Yang, Ruihan and Srivastava, Prakhar and Mandt, Stephan},
  journal={Entropy},
  volume={25},
  number={10},
  pages={1469},
  year={2023},
  publisher={MDPI}
}

@inproceedings{peebles2023scalable,
  title={Scalable diffusion models with transformers},
  author={Peebles, William and Xie, Saining},
  booktitle={Proceedings of the IEEE/CVF international conference on computer vision},
  pages={4195--4205},
  year={2023}
}

@article{jia2024bench2drive,
  title={Bench2drive: Towards multi-ability benchmarking of closed-loop end-to-end autonomous driving},
  author={Jia, Xiaosong and Yang, Zhenjie and Li, Qifeng and Zhang, Zhiyuan and Yan, Junchi},
  journal={Advances in Neural Information Processing Systems},
  volume={37},
  pages={819--844},
  year={2024}
}

@inproceedings{kim2023neuralfield,
  title={Neuralfield-ldm: Scene generation with hierarchical latent diffusion models},
  author={Kim, Seung Wook and Brown, Bradley and Yin, Kangxue and Kreis, Karsten and Schwarz, Katja and Li, Daiqing and Rombach, Robin and Torralba, Antonio and Fidler, Sanja},
  booktitle={Proceedings of the IEEE/CVF conference on computer vision and pattern recognition},
  pages={8496--8506},
  year={2023}
}

@inproceedings{yang2023law,
  title={Law-diffusion: Complex scene generation by diffusion with layouts},
  author={Yang, Binbin and Luo, Yi and Chen, Ziliang and Wang, Guangrun and Liang, Xiaodan and Lin, Liang},
  booktitle={Proceedings of the IEEE/CVF International Conference on Computer Vision},
  pages={22669--22679},
  year={2023}
}

@inproceedings{chitta2021neat,
  title={Neat: Neural attention fields for end-to-end autonomous driving},
  author={Chitta, Kashyap and Prakash, Aditya and Geiger, Andreas},
  booktitle={Proceedings of the IEEE/CVF International Conference on Computer Vision},
  pages={15793--15803},
  year={2021}
}

@inproceedings{liao2025diffusiondrive,
  title={Diffusiondrive: Truncated diffusion model for end-to-end autonomous driving},
  author={Liao, Bencheng and Chen, Shaoyu and Yin, Haoran and Jiang, Bo and Wang, Cheng and Yan, Sixu and Zhang, Xinbang and Li, Xiangyu and Zhang, Ying and Zhang, Qian and others},
  booktitle={Proceedings of the Computer Vision and Pattern Recognition Conference},
  pages={12037--12047},
  year={2025}
}

@article{li2025hydraplus,
  title={Hydra-mdp++: Advancing end-to-end driving via expert-guided hydra-distillation},
  author={Li, Kailin and Li, Zhenxin and Lan, Shiyi and Xie, Yuan and Zhang, Zhizhong and Liu, Jiayi and Wu, Zuxuan and Yu, Zhiding and Alvarez, Jose M},
  journal={arXiv preprint arXiv:2503.12820},
  year={2025}
}

@article{graphdiff,
  title={Graph-based interaction-aware multimodal 2D vehicle trajectory prediction using diffusion graph convolutional networks},
  author={Wu, Keshu and Zhou, Yang and Shi, Haotian and Li, Xiaopeng and Ran, Bin},
  journal={IEEE Transactions on Intelligent Vehicles},
  volume={9},
  number={2},
  pages={3630--3643},
  year={2023},
  publisher={IEEE}
}

@article{guo2025ipad,
  title={iPad: Iterative Proposal-centric End-to-End Autonomous Driving},
  author={Guo, Ke and Liu, Haochen and Wu, Xiaojun and Pan, Jia and Lv, Chen},
  journal={arXiv preprint arXiv:2505.15111},
  year={2025}
}

@article{li2025end,
  title={End-to-end driving with online trajectory evaluation via bev world model},
  author={Li, Yingyan and Wang, Yuqi and Liu, Yang and He, Jiawei and Fan, Lue and Zhang, Zhaoxiang},
  journal={arXiv preprint arXiv:2504.01941},
  year={2025}
}

@article{dauner2024navsim,
  title={Navsim: Data-driven non-reactive autonomous vehicle simulation and benchmarking},
  author={Dauner, Daniel and Hallgarten, Marcel and Li, Tianyu and Weng, Xinshuo and Huang, Zhiyu and Yang, Zetong and Li, Hongyang and Gilitschenski, Igor and Ivanovic, Boris and Pavone, Marco and others},
  journal={Advances in Neural Information Processing Systems},
  volume={37},
  pages={28706--28719},
  year={2024}
}

@inproceedings{ICLR2025_6aa49679,
 author = {Li, Yingyan and Fan, Lue and He, Jiawei and Wang, Yuqi and Chen, Yuntao and Zhang, Zhaoxiang and Tan, Tieniu},
 booktitle = {International Conference on Representation Learning},
 editor = {Y. Yue and A. Garg and N. Peng and F. Sha and R. Yu},
 pages = {42942--42959},
 title = {Enhancing End-to-End Autonomous Driving with Latent World Model},
 url = {https://proceedings.iclr.cc/paper_files/paper/2025/file/6aa4967920e495e90aeeaa3acf18d019-Paper-Conference.pdf},
 volume = {2025},
 year = {2025}
}

@article{bojarski2016end,
  title={End to end learning for self-driving cars},
  author={Bojarski, Mariusz and Del Testa, Davide and Dworakowski, Daniel and Firner, Bernhard and Flepp, Beat and Goyal, Prasoon and Jackel, Lawrence D and Monfort, Mathew and Muller, Urs and Zhang, Jiakai and others},
  journal={arXiv preprint arXiv:1604.07316},
  year={2016}
}

@inproceedings{peng2023openscene,
  title={Openscene: 3d scene understanding with open vocabularies},
  author={Peng, Songyou and Genova, Kyle and Jiang, Chiyu and Tagliasacchi, Andrea and Pollefeys, Marc and Funkhouser, Thomas and others},
  booktitle={Proceedings of the IEEE/CVF conference on computer vision and pattern recognition},
  pages={815--824},
  year={2023}
}

@inproceedings{dosovitskiy2017carla,
  title={CARLA: An open urban driving simulator},
  author={Dosovitskiy, Alexey and Ros, German and Codevilla, Felipe and Lopez, Antonio and Koltun, Vladlen},
  booktitle={Conference on robot learning},
  pages={1--16},
  year={2017},
  organization={PMLR}
}

@article{vaswani2017attention,
  title={Attention is all you need},
  author={Vaswani, Ashish and Shazeer, Noam and Parmar, Niki and Uszkoreit, Jakob and Jones, Llion and Gomez, Aidan N and Kaiser, {\L}ukasz and Polosukhin, Illia},
  journal={Advances in neural information processing systems},
  volume={30},
  year={2017}
}

@inproceedings{xing2025goalflow,
  title={Goalflow: Goal-driven flow matching for multimodal trajectories generation in end-to-end autonomous driving},
  author={Xing, Zebin and Zhang, Xingyu and Hu, Yang and Jiang, Bo and He, Tong and Zhang, Qian and Long, Xiaoxiao and Yin, Wei},
  booktitle={Proceedings of the Computer Vision and Pattern Recognition Conference},
  pages={1602--1611},
  year={2025}
}

@article{cao2025navsimv2,
  title={Pseudo-simulation for autonomous driving},
  author={Cao, Wei and Hallgarten, Marcel and Li, Tianyu and Dauner, Daniel and Gu, Xunjiang and Wang, Caojun and Miron, Yakov and Aiello, Marco and Li, Hongyang and Gilitschenski, Igor and others},
  journal={arXiv preprint arXiv:2506.04218},
  year={2025}
}

@article{ho2020denoising,
  title={Denoising diffusion probabilistic models},
  author={Ho, Jonathan and Jain, Ajay and Abbeel, Pieter},
  journal={Advances in neural information processing systems},
  volume={33},
  pages={6840--6851},
  year={2020}
}

@article{su2024difsd,
  title={Difsd: Ego-centric fully sparse paradigm with uncertainty denoising and iterative refinement for efficient end-to-end autonomous driving},
  author={Su, Haisheng and Wu, Wei and Yan, Junchi},
  journal={arXiv e-prints},
  pages={arXiv--2409},
  year={2024}
}

@inproceedings{rombach2022high,
  title={High-resolution image synthesis with latent diffusion models},
  author={Rombach, Robin and Blattmann, Andreas and Lorenz, Dominik and Esser, Patrick and Ommer, Bj{\"o}rn},
  booktitle={Proceedings of the IEEE/CVF conference on computer vision and pattern recognition},
  pages={10684--10695},
  year={2022}
}

@article{janner2022planning,
  title={Planning with diffusion for flexible behavior synthesis},
  author={Janner, Michael and Du, Yilun and Tenenbaum, Joshua B and Levine, Sergey},
  journal={arXiv preprint arXiv:2205.09991},
  year={2022}
}

@article{kondo2024cgd,
  title={Cgd: Constraint-guided diffusion policies for uav trajectory planning},
  author={Kondo, Kota and Tagliabue, Andrea and Cai, Xiaoyi and Tewari, Claudius and Garcia, Olivia and Espitia-Alvarez, Marcos and How, Jonathan P},
  journal={arXiv preprint arXiv:2405.01758},
  year={2024}
}

@article{zheng2025diffusion,
  title={Diffusion-based planning for autonomous driving with flexible guidance},
  author={Zheng, Yinan and Liang, Ruiming and Zheng, Kexin and Zheng, Jinliang and Mao, Liyuan and Li, Jianxiong and Gu, Weihao and Ai, Rui and Li, Shengbo Eben and Zhan, Xianyuan and others},
  journal={arXiv preprint arXiv:2501.15564},
  year={2025}
}

@article{zhu2023diffusion,
  title={Diffusion models for reinforcement learning: A survey},
  author={Zhu, Zhengbang and Zhao, Hanye and He, Haoran and Zhong, Yichao and Zhang, Shenyu and Guo, Haoquan and Chen, Tingting and Zhang, Weinan},
  journal={arXiv preprint arXiv:2311.01223},
  year={2023}
}

@inproceedings{wang2024trajfine,
  title={TrajFine: Predicted trajectory refinement for pedestrian trajectory forecasting},
  author={Wang, Kuan-Lin and Tsao, Li-Wu and Wu, Jhih-Ciang and Shuai, Hong-Han and Cheng, Wen-Huang},
  booktitle={Proceedings of the IEEE/CVF Conference on Computer Vision and Pattern Recognition},
  pages={4483--4492},
  year={2024}
}

@inproceedings{bae2024singulartrajectory,
  title={Singulartrajectory: Universal trajectory predictor using diffusion model},
  author={Bae, Inhwan and Park, Young-Jae and Jeon, Hae-Gon},
  booktitle={Proceedings of the IEEE/CVF Conference on Computer Vision and Pattern Recognition},
  pages={17890--17901},
  year={2024}
}

\appendix
\setcounter{secnumdepth}{2}

\maketitle

\section*{Appendix}   
\addcontentsline{toc}{section}{Appendix} 

\section{Problem Definition and Formulation}
\label{app:problem}

End-to-end autonomous driving aims to directly predict the future trajectory of the ego vehicle from raw sensor inputs. The predicted trajectory is represented as \(\tau = \{\tau_i\}_{i=1}^T\), where \(T\) denotes the prediction horizon and \(\tau_i\) corresponds to the ego vehicle's state at time step \(i\), including its position and heading.

Inspired by DiffusionDrive~\cite{liao2025diffusiondrive}, we propose a two-stage anchor-conditioned diffusion framework for trajectory generation.

\paragraph{Proposal-based Diffusion.}  
Our method decomposes the clean trajectory \(\tau^0\) into three components: a deterministic anchor \(a_k\) obtained via offline clustering, a learnable offset \(\delta_k\), and Gaussian noise \(\epsilon\) introduced by the diffusion process.

\paragraph{Proposal Construction.}  
The clean trajectory is formulated as:
\begin{equation}
\tau^0 = a_k + \delta_k,
\end{equation}
where \(a_k\) is the \(k\)-th anchor center and \(\delta_k\) is the offset from the anchor to the ground-truth trajectory, predicted by an offset network. Denoting the predicted offset as \(\hat{\delta}_k\), the estimated clean trajectory is:
\begin{equation}
\hat{\tau}^0 = a_k + \hat{\delta}_k.
\end{equation}

\begin{algorithm}[t]
\caption{DiffRefiner Pipeline with Semantic Interaction}
\label{alg:diffrefiner}
\begin{algorithmic}[1]
\State \textbf{Input:} Multi-view camera images $\mathcal{I}$, ego vehicle status $\mathbf{E}$
\State \textbf{Output:} Final predicted trajectory $\hat{\tau}$

\State Extract BEV feature $\mathbf{F}_{\text{bev}}$ from $\mathcal{I}$ using a shared perception backbone

\State Perform semantic segmentation on $\mathbf{F}_{\text{bev}}$ to obtain semantic map $\mathbf{S}$
\State Perform object detection based on $\mathbf{F}_{\text{bev}}$ and ego status $\mathbf{E}$ to obtain detection results $\mathbf{D}$

\State Load pre-defined anchor set $\{\mathbf{a}_k\}_{k=1}^K$ from offline clustering

\State \textbf{[In parallel for each anchor $\mathbf{a}_k$]:}
\State \quad Use proposal decoder to predict offset $\delta_k$ from $(\mathbf{a}_k, \mathbf{F}_{\text{bev}}, \mathbf{E})$
\State \quad Compute initial trajectory $\tau_k^0 = \mathbf{a}_k + \delta_k$

\Statex \Comment{\textit{Training only: add Gaussian noise to enable diffusion denoising}}
\If{training}
    \State \textbf{[In parallel for each trajectory $\tau_k^0$]:}
    \State \quad Add noise: $\tau_k^T = \tau_k^0 + \epsilon$, where $\epsilon \sim \mathcal{N}(0, \sigma^2)$
\Else
    \State Set $\tau_k^T = \tau_k^0$
\EndIf

\State \textbf{[In parallel for each trajectory $\tau_k^T$]:}
\For{$t = T$ \textbf{to} $1$}
    \State Fuse $\tau_k^t$, $\mathbf{S}$, $\mathbf{F}_{\text{bev}}$, and $\mathbf{E}$ via FGSIM
    \State Predict residual noise $\hat{\epsilon}_t$ using diffusion decoder
    \State Apply DDIM update to obtain $\tau_k^{t-1}$
\EndFor

\State Compute confidence scores $\{s_k\}_{k=1}^K$ for all refined trajectories $\{\tau_k^0\}$
\State Select final output: $\hat{\tau} = \arg\max_k s_k$
\end{algorithmic}
\end{algorithm}

\paragraph{Forward Diffusion.}  
Differing from DiffusionDrive~\cite{liao2025diffusiondrive}, which applies diffusion directly from the anchor, we perform forward diffusion~\cite{ho2020denoising} based on the offset-corrected anchor. Given the estimate \(\hat{\tau}^0\), the forward diffusion process at step \(i\) is expressed as:
\begin{equation}
\tau^i = \sqrt{\bar{\alpha}^i} \, \hat{\tau}^0 + \sqrt{1 - \bar{\alpha}^i} \, \epsilon, \quad \epsilon \sim \mathcal{N}(0, \mathbf{I}).
\end{equation}

\paragraph{Reverse Denoising.}  
To recover the clean trajectory \(\tau^0\) from the noisy observation \(\tau^i\), we train a denoising network \(\epsilon_\theta\) to predict the noise component \(\epsilon\):
\begin{equation}
\hat{\epsilon}_i = \epsilon_\theta(\tau^i, i).
\end{equation}
The trajectory at step \(i-1\) is then reconstructed via the DDIM~\cite{song2021denoising} update:
\begin{equation}
\hat{\tau}^{i-1} = \frac{1}{\sqrt{\bar{\alpha}^i}} \left( \tau^i - \sqrt{1 - \bar{\alpha}^i} \, \hat{\epsilon}_i \right).
\end{equation}

\noindent
By diffusing from the offset-corrected anchor rather than the anchor alone, our approach requires fewer denoising steps, enhancing computational efficiency and meeting the real-time demands of autonomous driving systems. 

For a detailed description of the algorithmic procedure, please refer to Algorithm~\ref{alg:diffrefiner}.

\section{Evaluation Metrics}
\paragraph{NAVSIM} The official NAVSIM v1 benchmark~\cite{dauner2024navsim} comprises five sub-metrics: No-at-fault Collision (\(NC\)), Drivable Area Compliance (\(DAC\)), Time-to-Collision (\(TTC\)), Ego Vehicle Progress (\(EP\)), and History Comfort (\(C\), referred to as \(HC\) in NAVSIM v2). 
The overall Predictive Driver Model Score (\(PDMS\)) is computed as:
\begin{equation}
PDMS = NC \times DAC \times \frac{5 \times TTC + 2 \times C + 5 \times EP}{12}.
\end{equation}

The NAVSIM v2 benchmark~\cite{cao2025navsimv2} extends the evaluation protocol by introducing additional sub-metrics, including Driving Direction Compliance (\(DDC\)), Traffic Light Compliance (\(TLC\)), Lane Keeping (\(LK\)), and Extended Comfort (\(EC\)), combined with a False-Positive Penalty Filtering mechanism. These additions provide a more comprehensive assessment of driving behavior. The final Extended Predictive Driver Model Score (\(\mathrm{EPDMS}\)) is computed as follows:

\begin{equation}
P = NC \times DAC \times DDC \times TLC
\end{equation}

\begin{equation}
\bar{M} = \frac{\sum_{m \in \{TTC, EP, HC, LK, EC\}} w_m m}{\sum_{m \in \{TTC, EP, HC, LK, EC\}} w_m}
\end{equation}

\begin{equation}
\mathrm{EPDMS} = P \times \bar{M}
\end{equation}

\noindent
where \(P\) denotes the product of core compliance metrics, which collectively reflect adherence to fundamental driving rules. \(\bar{M}\) represents the weighted average of supplementary driving behavior indicators, with weights \(w_m\) encoding the relative importance of each metric. Specifically, the weights are set as follows: \(w_{TTC} = 5\), \(w_{EP} = 5\), \(w_{HC} = 2\), \(w_{LK} = 2\), and \(w_{EC} = 2\). The final score \(\mathrm{EPDMS}\) integrates these components to provide a holistic evaluation of driving performance.

\begin{table*}
    \centering
    \fontsize{9pt}{9pt}\selectfont
    \setlength{\tabcolsep}{25pt}  
    \begin{tabular}{c|l|c|c}
        \toprule
        Type & Parameter & Symbol & Value \\
        \midrule
        \multirow{8}{*}{Loss Weight} 
            & segmentation loss of perception & $w_{\mathrm{seg}}$ & 14.0 \\
            & classification loss of perception & $w_{\mathrm{type}}$ & 10.0 \\
            & bounding box regression loss of perception & $w_{\mathrm{box}}$ & 1.0 \\
            & classification loss of planning & $w_{\mathrm{cls}}$ & 10.0 \\
            & regression loss of planning & $w_{\mathrm{reg}}$ & 8.0 \\
            & proposal loss (total) & $w_{\mathrm{pro}}$ & 12.0 \\
            & refinement loss (total) & $w_{\mathrm{ref}}$ & 12.0 \\
            & perception loss (total) & $w_{\mathrm{perc}}$ & 1.0 \\
        \midrule
        \multirow{3}{*}{Training Process} 
            & batch size & $B$ & 384 \\
            & learning rate & $\eta$ & $4\times10^{-4}$ \\
            & weight decay & $\lambda_{\mathrm{wd}}$ & $1\times10^{-4}$ \\
        \bottomrule
    \end{tabular}
    \caption{Hyperparameters and corresponding symbols.}
    \label{tab:Hyperparameters_navsim}
\end{table*}

\paragraph{Bench2Drive} 
For the Bench2Drive benchmark~\cite{jia2024bench2drive}, we adopt CARLA’s official Driving Score (DS) and Success Rate (SR) metrics~\cite{dosovitskiy2017carla}. The DS jointly evaluates the agent’s route completion and penalties incurred from traffic infractions, providing a comprehensive measure of driving effectiveness. The SR reflects the proportion of test scenarios in which the agent successfully completes the entire route without any violations. Beyond DS and SR, we supplement the evaluation with multi-ability metrics to assess the model’s performance across diverse driving scenarios and tasks, including merging, overtaking, yielding, traffic sign compliance, and emergency braking.

\begin{table*}
    \centering
    \fontsize{9pt}{9pt}\selectfont
    \setlength{\tabcolsep}{25pt}  
    \begin{tabular}{c|l|c|c}
        \toprule
        Type & Parameter & Symbol & Value \\
        \midrule
        \multirow{8}{*}{Loss Weight} 
            & image segmentation loss of perception & $w_{\mathrm{seg_img}}$ & 1.0 \\
            & image depth loss of perception & $w_{\mathrm{depth}}$ & 1.0 \\
            & BEV segmentation loss of perception & $w_{\mathrm{seg_bev}}$ & 1.0 \\
            & classification loss of perception & $w_{\mathrm{type}}$ & 1.0 \\
            & bounding box regression loss of perception & $w_{\mathrm{box}}$ & 1.0 \\
            & agent prediction loss & $w_{\mathrm{agent}}$ & 1.0 \\
            & speed loss & $w_{\mathrm{speed}}$ & 1.0 \\
            & classification loss of planning & $w_{\mathrm{cls}}$ & 10.0 \\
            & regression loss of planning & $w_{\mathrm{reg}}$ & 8.0 \\
            & proposal loss (total) & $w_{\mathrm{pro}}$ & 1.0 \\
            & refinement loss (total) & $w_{\mathrm{ref}}$ & 1.0 \\
        \midrule
        \multirow{4}{*}{Training Process} 
            & batch size (stage 1) & $B_{\mathrm{stage1}}$ & 128 \\
            & batch size (stage 2) & $B_{\mathrm{stage2}}$ & 512 \\
            & learning rate & $\eta$ & $3\times10^{-4}$ \\
            & weight decay & $\lambda_{\mathrm{wd}}$ & $1\times10^{-2}$ \\
        \bottomrule
    \end{tabular}
    \label{tab:Hyperparameters_carla}
    \caption{Hyperparameters and corresponding symbols.}
\end{table*}

\begin{table*}[htbp]
    \centering
    \renewcommand{\arraystretch}{1.1}
    \setlength{\tabcolsep}{5pt}
    \begin{tabular}{c|c|c|c|cc|ccc}
        \toprule
        Method & Backbone & Modality & PDMS$\uparrow$ & NC$\uparrow$ & DAC$\uparrow$ &TTC$\uparrow$ & EP$\uparrow$ & C$\uparrow$ \\
        \midrule
        Human Agent & - & - & 94.8 & 100 & 100 & 100 & 87.5 & 99.9 \\
        \midrule
        Transfuser~\cite{jaeger2023tf++} & ResNet34 & C+L & 84.0 &  97.7 & 92.8  & 92.8  & 79.2  & 100    \\
        DiffusionDrive~\cite{liao2025diffusiondrive} & ResNet34 & C+L & 88.1  & 98.2  & 96.2  & 94.7  & 82.2  & 100  \\
        WoTE~\cite{li2025end} & ResNet34 & C+L &  88.3 & \textbf{98.5}  &  96.8 & 94.9 & 81.9  & 99.9  \\
        GaussianFusion~\cite{liu2025gaussianfusion} & ResNet34 & C+L & 88.8  & 98.3  & 97.2  & 94.6  & 83.0  & -  \\
        UniAD~\cite{uniad} & ResNet34 & C &  83.4 & 97.8  & 91.9  & 92.9  & 78.8  & 100  \\
        PARA-Drive~\cite{weng2024paradrive} & ResNet34& C & 84.0  & 97.9  & 92.4  &  93.0 & 79.3  & 99.8  \\
        VADv2~\cite{chen2024vadv2} & ResNet34 & C & 83.0  & 97.9  & 91.7  & 92.9  & 77.6  & 100  \\
        Hydra-MDP~\cite{li2024hydra} & ResNet34 & C & 86.5  & 98.3  & 96.0  & 94.6  & 78.7  & 100  \\
        \textbf{DiffRefiner} & ResNet34 & C & \textbf{89.4} &  98.4 &  \textbf{97.4} & \textbf{95.3}  & \textbf{83.4}  & \textbf{100}  \\
        \midrule
        GoalFlow~\cite{xing2025goalflow} & V2-99 & C+L & 90.3  & 98.4  & 98.3  & 94.6  & \textbf{85.0}  & 100  \\
        \textbf{DiffRefiner} & V2-99 & C &  \textbf{90.7} &  \textbf{98.6} &  \textbf{98.4} &  \textbf{95.8} &  84.5 & \textbf{100}  \\
        \bottomrule
    \end{tabular}
    \caption{Evaluation on NAVSIM v1. Results are grouped by backbone types.}
    \label{tab:results-navsimv1}
\end{table*}

\section{Model Details}
\paragraph{NAVSIM} In our NAVSIM experiments, we predict vehicle trajectories for the next 4 seconds at 0.5-second intervals, resulting in 8 future trajectory points. We follow the base settings of DiffusionDrive~\cite{liao2025diffusiondrive}, but remove LiDAR inputs and use only camera data.
The perception network is designed following the Transfuser-based architecture and is supervised by three components: a semantic segmentation classification loss $\mathcal{L}_{\mathrm{seg}}$, an object classification loss $\mathcal{L}_{\mathrm{type}}$, and a bounding box regression loss $\mathcal{L}_{\mathrm{box}}$. 
The overall perception loss is defined as:
\begin{equation}
\mathcal{L}_{\mathrm{perception}} = 
w_{\mathrm{seg}}\mathcal{L}_{\mathrm{seg}} +
w_{\mathrm{type}}\mathcal{L}_{\mathrm{type}} +
w_{\mathrm{box}}\mathcal{L}_{\mathrm{box}},
\end{equation}
where $w_{\mathrm{seg}}$, $w_{\mathrm{cls}}$, and $w_{\mathrm{box}}$ are the corresponding weighting coefficients.
For the planning task, both the proposal and refinement modules contain classification and regression branches.
A winner-takes-all strategy selects the trajectory candidate $\tau^*$ closest to ground truth $\tau^{gt}$, with losses computed as:
\begin{equation}
\mathcal{L}_{\mathrm{planning}} =
w_{\mathrm{cls}}\mathcal{L}_{\mathrm{cls}} +
w_{\mathrm{reg}}\mathcal{L}_{\mathrm{reg}}
\end{equation}

\begin{equation}
\mathcal{L}_{\mathrm{reg}} = |\tau^* - \tau^{gt}|_1
\end{equation}

\begin{equation}
\mathcal{L}_{\mathrm{cls}} = -\sum_{i=1}^N \left[
\mathbb{I}(i = *) \log p_i +
\mathbb{I}(i \neq *) \log(1 - p_i)
\right]
\end{equation}

where:
\begin{itemize}
\item $\tau^*$ is the selected trajectory candidate closest to the ground truth trajectory
\item $\tau_{gt}$ is the ground truth trajectory
\item $p_i$ is the predicted probability for the $i$-th candidate
\item $\mathbb{I}(\cdot)$ is the indicator function
\item $N$ is the total number of trajectory candidates
\end{itemize}

The overall training objective combines the proposal and refinement losses as:
\begin{equation}
\mathcal{L}_{\mathrm{total}} =
w_{\mathrm{pro}}\mathcal{L}_{\mathrm{proposal}} +
w_{\mathrm{ref}}\mathcal{L}_{\mathrm{refine}} +
w_{\mathrm{perc}}\mathcal{L}_{\mathrm{perception}}.
\end{equation}

We adopt a two-stage training paradigm: the first stage trains the perception network, while the second stage jointly optimizes perception and planning in an end-to-end manner.
For trajectory anchors, we follow DiffusionDrive~\cite{liao2025diffusiondrive} and use a total of 20 anchors. In the first stage, the proposal decoder adjusts the clustered anchors to generate adjusted anchors, serving as coarse trajectories and providing priors for the second-stage refinement module.
Although the first stage can optionally perform a simple scoring task, empirical results show that passing all 20 adjusted trajectories from the first stage into the refinement process yields better performance.
During training, we set the forward diffusion process to 50 steps, while at inference time we directly utilize the proposal trajectories without adding noise.

For the refinement stage, we incorporate two instances of the Fine-Grained Semantic Interaction Module(FGSIM), sequentially enhancing interactions with the drivable area and traffic participants. In each instance, the target regions are extracted using semantic masks corresponding to specific labels. The drivable area mask includes both road surfaces and centerlines, while the traffic participant mask covers vehicles and pedestrians.

\paragraph{Bench2Drive}
In the closed-loop CARLA experiments, we follow the base configuration of TF++~\cite{zimmerlin2024tf++dataset}, using both LiDAR and camera sensors as inputs. 
The perception network is trained with multiple auxiliary tasks, including image semantic segmentation, camera depth estimation, BEV semantic segmentation, and agent prediction. 
The corresponding losses are denoted as 
$\mathcal{L}_{\mathrm{img\_seg}}$, 
$\mathcal{L}_{\mathrm{depth}}$, 
$\mathcal{L}_{\mathrm{bev\_seg}}$, and 
$\mathcal{L}_{\mathrm{agent}}$. 

We adopt the same training dataset as TF++~\cite{zimmerlin2024tf++dataset}. Since TF++ decouples trajectory generation into speed prediction and path prediction, we retain its original speed prediction branch with a loss denoted as $\mathcal{L}_{\mathrm{speed}}$, while enhancing path quality through a two-stage planning network that predicts 10 future waypoints spanning approximately 10 meters.
The planning loss follows the same formulation as in the NAVSIM experiments.

In the first stage, we pretrain the perception network. 
To save GPU memory, during the second stage we freeze the perception network and train only the planning module. 
The stage-specific objectives are formulated as:
\begin{equation}
\mathcal{L}_{\mathrm{stage1}} = 
\mathcal{L}_{\mathrm{perception}},
\end{equation}
\begin{equation}
\mathcal{L}_{\mathrm{stage2}} = 
\mathcal{L}_{\mathrm{proposal}} + \mathcal{L}_{\mathrm{refinement}} + \mathcal{L}_{\mathrm{speed}},
\end{equation}
where $\mathcal{L}_{\mathrm{perception}}$ denotes the combined perception losses and $\mathcal{L}_{\mathrm{proposal}}$ and $\mathcal{L}_{\mathrm{refinement}}$ follows the NAVSIM definition.

For the Fine-Grained Semantic Interaction Module (FGSIM), we adopt a design similar to the NAVSIM experiments.

\begin{figure*}[htbp]
\centering
\includegraphics[width=1.0\linewidth]{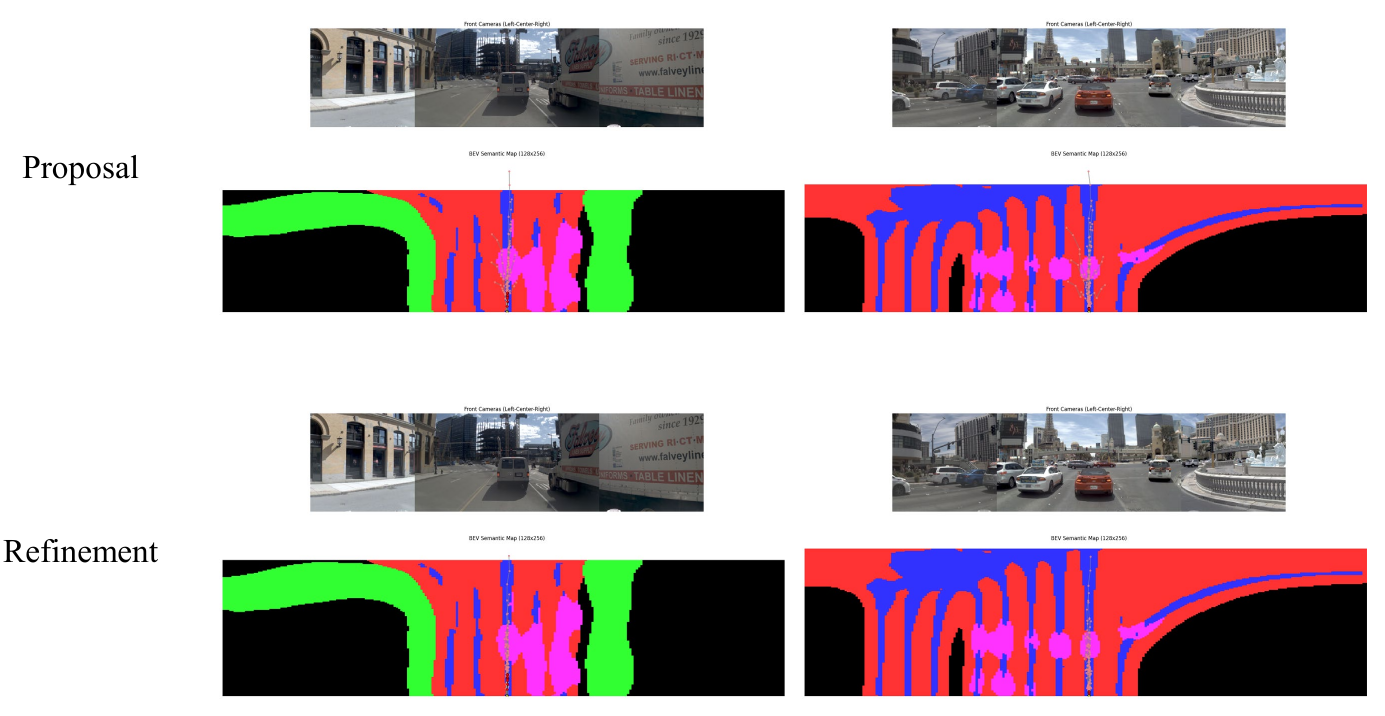}
\caption{Qualitative examples of the proposal refinement process on NAVSIM. 
Red regions indicate roads, blue represents centerlines, pink denotes vehicles, and green corresponds to walkways. 
The ego vehicle is positioned at the center of the bottom area. 
Gray lines depict all 20 predicted candidate trajectories.}

\label{fig:qual_1}
\end{figure*}

\section{Additional Experimental Results}
\paragraph{Performance On NAVSIM v1.} For a more comprehensive evaluation and comparison, we additionally validate our approach on the NAVSIM v1 benchmark. The model configuration is kept consistent with the NAVSIM v2 experiments, relying solely on camera sensors for perception. 

As shown in Table ~\ref{tab:results-navsimv1}, our method achieves improvements of 1.3 in PDMS and 1.2 in DAC compared to DiffusionDrive~\cite{liao2025diffusiondrive} when using the same ResNet34 backbone, demonstrating the superior scene alignment capability of our model. With the stronger V2-99 backbone, our approach attains a PDMS of 90.7, NC of 98.6, and DAC of 98.4, significantly outperforming existing methods and achieving state-of-the-art performance.

\begin{table}
    \centering
    \fontsize{9pt}{9pt}\selectfont
    \setlength{\tabcolsep}{15pt}  
    \begin{tabular}{c|cc|cc}
        \toprule
        ID & w/ Ref & w/ Pro & DS$\uparrow$  & SR$\uparrow$\\
        \midrule
        1 &   & \checkmark   &  85.3  & 65.5 \\ 
        2 & \checkmark &       &  71.8 & 45.5 \\ 
        3 & \checkmark & \checkmark  & \textbf{91.0}   & \textbf{78.2} \\ 
        \bottomrule
    \end{tabular}
    \caption{Ablation study of the proposed planning framework under the closed-loop setting. 
    ``Ref'' and ``Pro'' denote the refinement and proposal stages, respectively. 
    Experiments are conducted on the \textit{Bench2Drive55} benchmark following \textit{HiPAD}~\cite{tang2025hip}.}
    \label{tab:ablation of framework}
\end{table}

\begin{table}[t]
    \centering
    \fontsize{9pt}{11pt}\selectfont
    \setlength{\tabcolsep}{8pt}
    \begin{tabular}{c|c|cccc}
        \toprule
        Top-$k$ & EPDMS$\uparrow$  & NC$\uparrow$  & DAC$\uparrow$ & DDC$\uparrow$ & TTC$\uparrow$ \\
        \midrule
        1   & 86.15 & 98.44 & 97.34 & 99.64 & 97.72 \\
        5   & 86.16 & 98.45 & 97.32 & 99.63 & 97.77 \\
        10  & 86.16 & 98.44 & 97.34 & 99.63 & 97.71 \\
        15  & 86.16 & 98.45 & 97.32 & 99.64 & 97.76 \\
        20  & \textbf{86.19} & \textbf{98.48} & \textbf{97.35} & \textbf{99.64} & \textbf{97.78} \\
        \bottomrule
    \end{tabular}
    \caption{Ablation study on the number of top-$k$ proposals forwarded to the refinement stage. 
    Increasing $k$ leads to marginal improvements, with the best performance achieved when all proposals are retained.}
    \label{tab:ablation_topk}
\end{table}

\paragraph{Ablation Study of Planning Framework in Bench2Drive.} 
To further validate the effectiveness of our proposed two-stage planning framework, we conduct ablation studies under the closed-loop setting of the Bench2Drive benchmark, as shown in Table~\ref{tab:ablation of framework}. The results demonstrate that the two-stage planning framework effectively enhances closed-loop performance and improves the model’s ability to capture fine-grained scene details.

\paragraph{Ablation Study on the Number of Top-$k$ Proposals.}
In the proposal stage, we introduce a classification head to rank candidate proposals. 
We conduct ablation studies to compare different top-$k$ selections that are passed into the subsequent refinement stage, as shown in Table~\ref{tab:ablation_topk}. 
The results indicate that forwarding all proposals to the refinement stage yields higher-quality final trajectories. 
This can be attributed to the lightweight nature of the first-stage proposal decoder, which may not be sufficiently expressive to reliably identify the best proposals at this stage.


\section{More Qualitative Results}  

\subsection{Proposal Refinement on NAVSIM}

Figure~\ref{fig:qual_1} illustrates the proposal refinement process on NAVSIM, demonstrating how coarse proposals are progressively optimized within our two-stage framework. The visualization shows that the initial proposals produce scene-adaptive candidates but often contain unsafe details. After refinement, the trajectories better align with the scene, reduce potential conflicts with surrounding agents, and achieve improved consistency with the map, thereby enhancing overall driving safety.

\subsection{More Qualitative Results on NAVSIM} 

\begin{figure*}[htbp]
\centering
\includegraphics[width=1.0\linewidth]{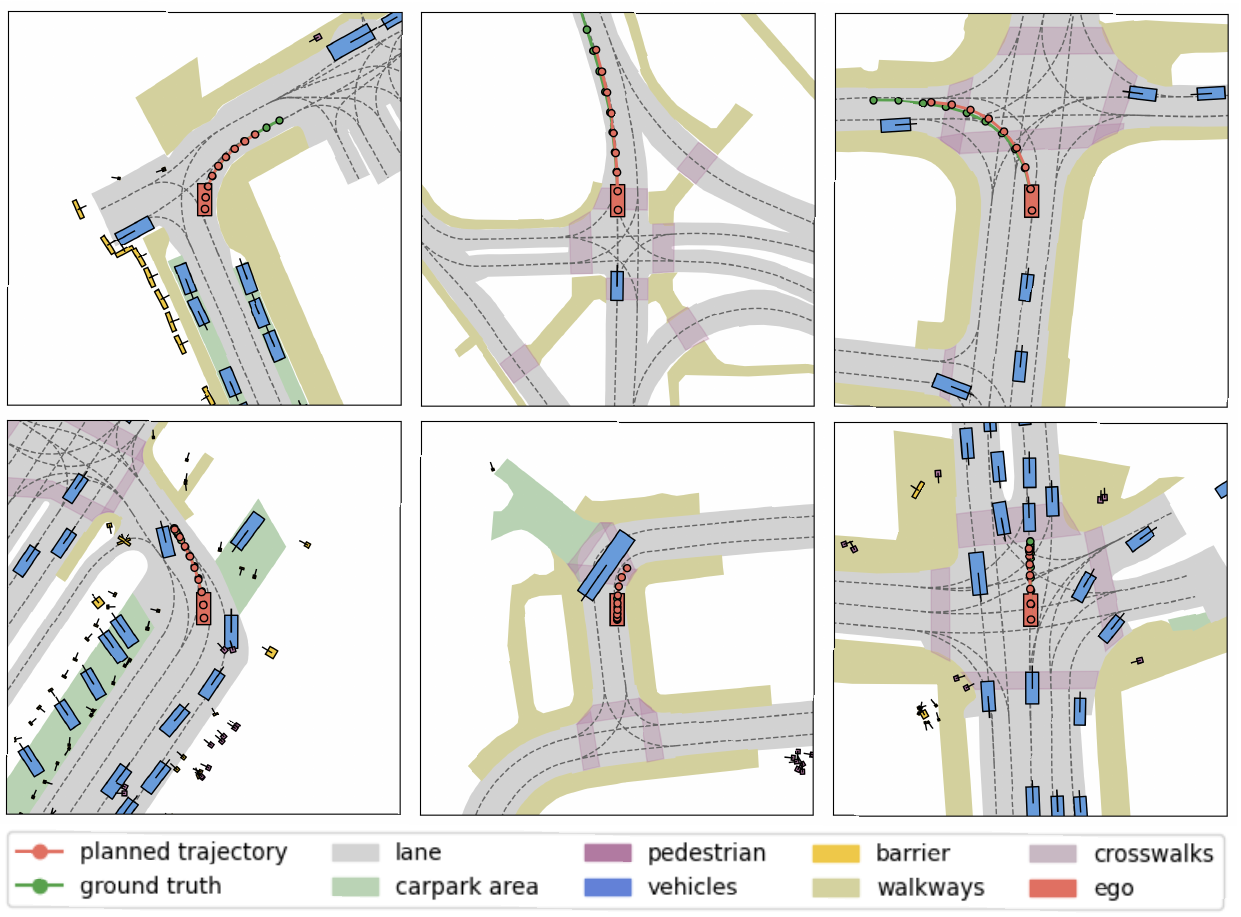}
\caption{Qualitative examples of trajectory prediction on NAVSIM under diverse and complex driving scenarios. 
The visualizations demonstrate the model's ability to generate scene-consistent trajectories that adapt to dynamic environments, interact safely with surrounding agents, and maintain strong alignment with the underlying map structure.}

\label{fig:qual_2}
\end{figure*}
Figure~\ref{fig:qual_2} presents additional qualitative results of our model on NAVSIM across diverse and complex driving scenarios, including dense urban environments and challenging interactive traffic situations. The visualizations demonstrate that our model is capable of generating scene-consistent and socially compliant trajectories, closely adhering to lane boundaries while navigating through intersections, curves, and merges. Moreover, the model exhibits strong interaction awareness by proactively adjusting its motion in response to surrounding obstacles, such as static roadblocks and dynamic agents, thereby maintaining safe and efficient driving behavior. These examples illustrate the model’s effectiveness in integrating lane-level map information and semantic context to make reliable decisions in multi-agent environments.

\subsection{More Qualitative Results on Bench2Drive} 

\begin{figure*}[htbp]
\centering
\includegraphics[width=1.0\linewidth]{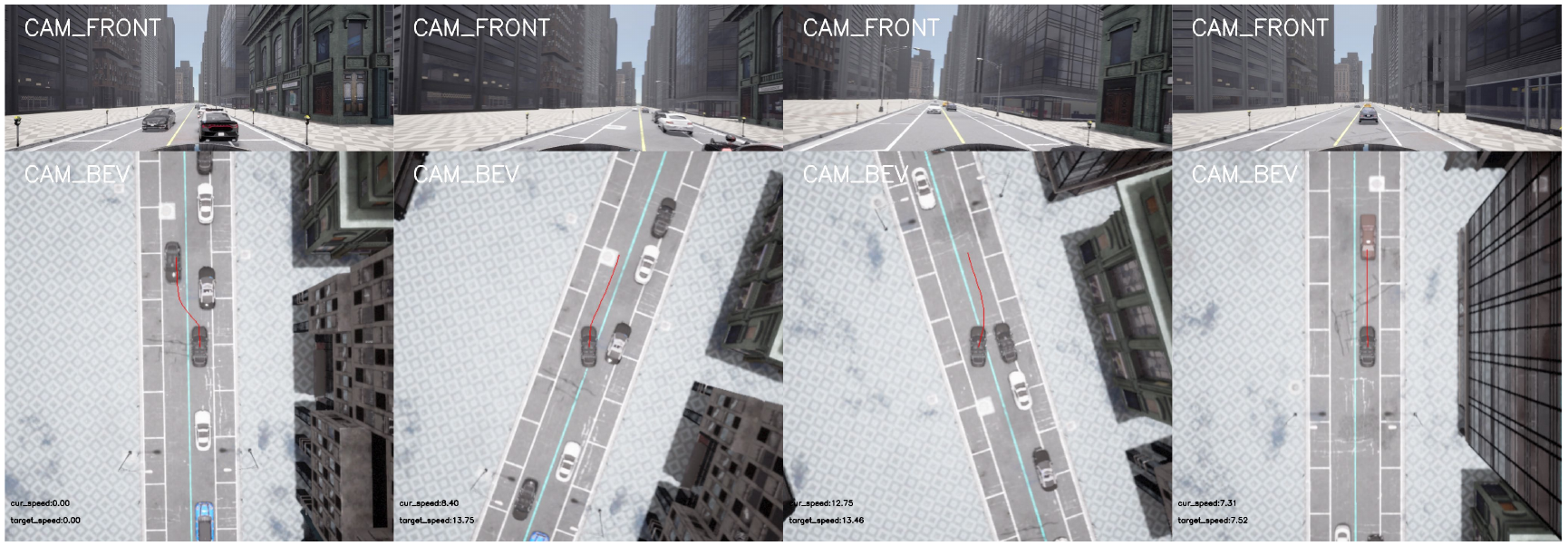}
\caption{Closed-loop qualitative case on CARLA, showing urban driving behaviors including stopping, waiting, detouring, and lane returning.}
\label{fig:qual_3}
\end{figure*}

Figure~\ref{fig:qual_3} present qualitative results on the CARLA closed-loop benchmark, demonstrating the model’s capability to manage urban driving tasks such as stopping, waiting, executing safe detours, and returning to the original lane.

\begin{figure*}[htbp]
\centering
\includegraphics[width=1.0\linewidth]{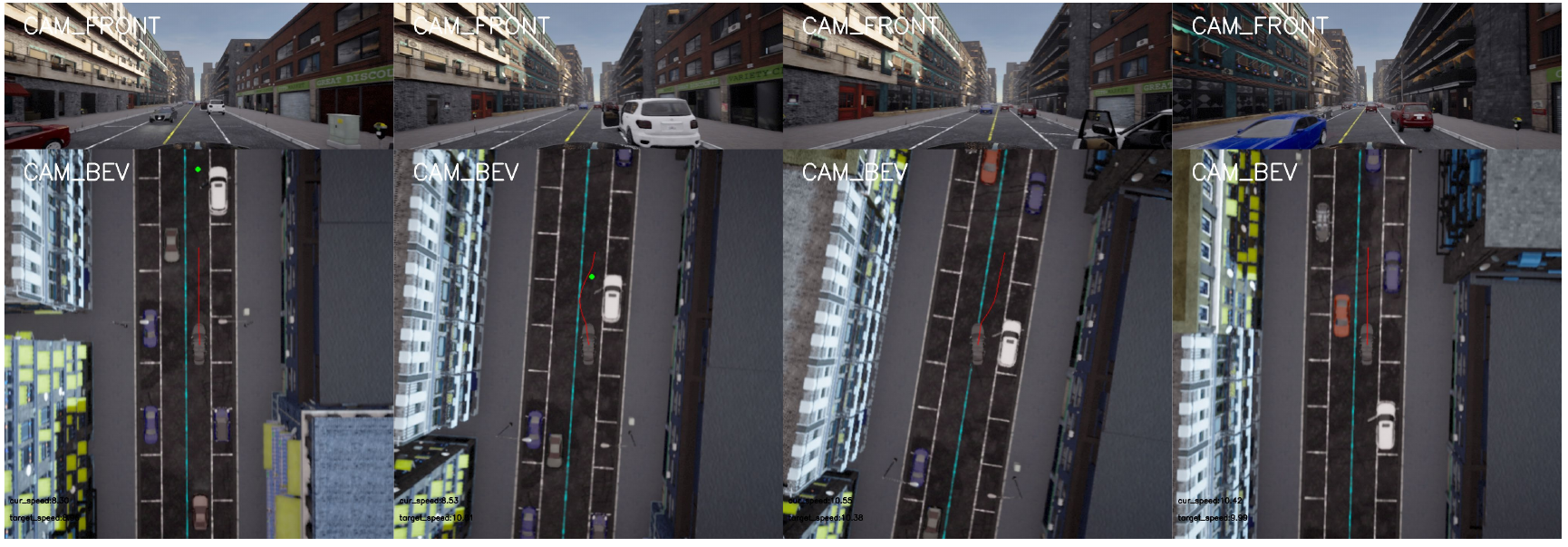}
\caption{Closed-loop qualitative case on CARLA, illustrating a corner-case scenario where the ego vehicle encounters a suddenly opened car door, performs a safe avoidance maneuver, detours around the obstacle, and returns to the original lane.}
\label{fig:qual_4}
\end{figure*}

Figure~\ref{fig:qual_4} illustrates a corner-case scenario where the model detects a suddenly opened car door, performs a safe avoidance maneuver, detours around the obstacle, and smoothly merges back into the original lane.

\begin{figure*}[htbp]
\centering
\includegraphics[width=0.95\linewidth]{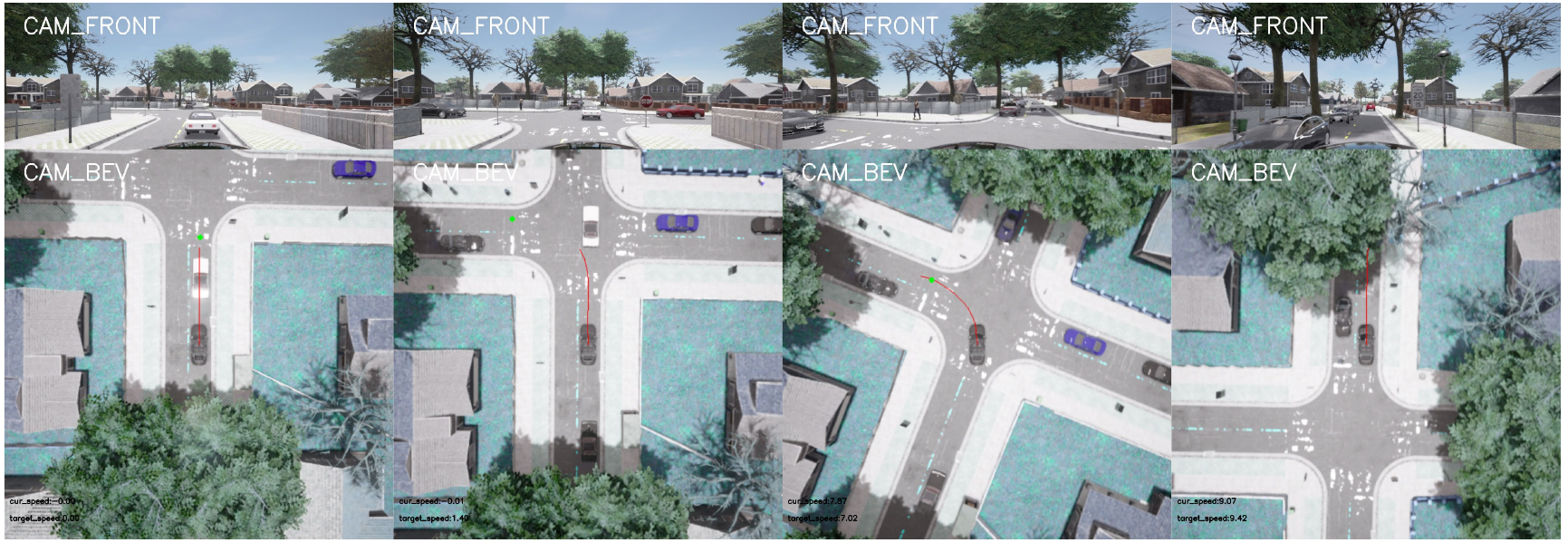}
\caption{Closed-loop qualitative case on CARLA, showing an intersection scenario where the ego vehicle stops to wait for the leading car, then performs a decisive left turn and proceeds smoothly into the target lane.}
\label{fig:qual_5}
\end{figure*}

Figure~\ref{fig:qual_5} showcases the model’s ability to handle intersection scenarios by stopping behind a leading vehicle, waiting safely, and executing a decisive left turn to merge smoothly into the designated lane once the path is clear.

\subsection{Failure Case Analysis}

To further investigate the limitations of our method, we analyze representative failure cases from the NAVSIM and CARLA benchmarks.

\subsubsection{NAVSIM Failure Cases}
\begin{figure*}[htbp]
\centering
\includegraphics[width=0.95\linewidth]{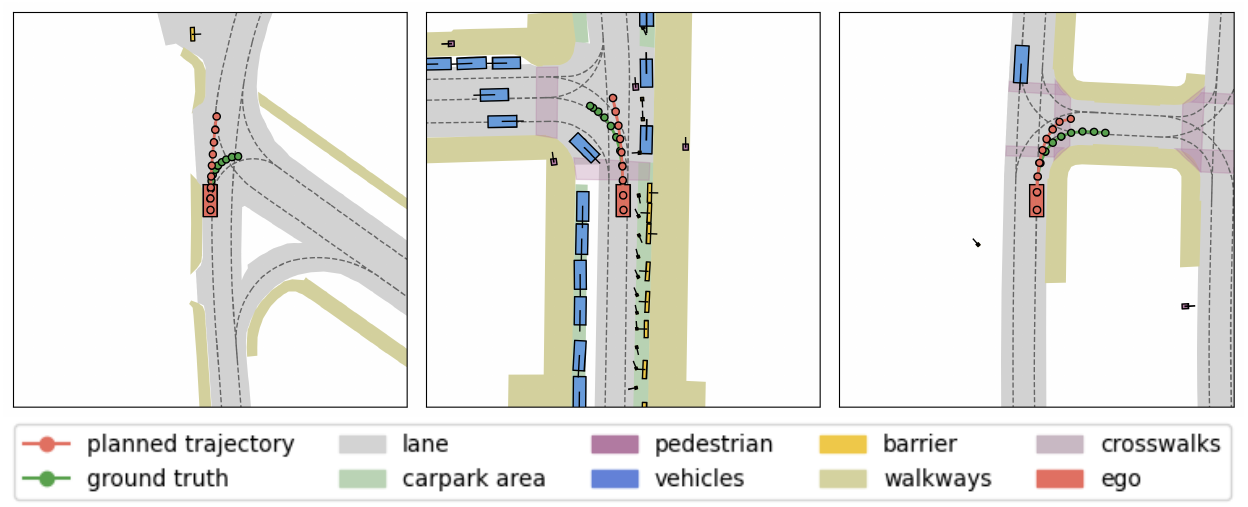}
\caption{Representative failure cases on NAVSIM.}

\label{fig:qual_6}
\end{figure*}

On the NAVSIM benchmark, as shown in Figure~\ref{fig:qual_6}, we observe three representative failure cases. One typical case involves the ego vehicle failing to follow the high-level navigation command, resulting in an incorrect turning decision. In the other two cases, although the vehicle attempts to perform a turning maneuver, it fails to select the correct target lane, leading to potentially dangerous driving behaviors. These cases reflect the model’s limitations in accurately interpreting navigation instructions and making precise lane-level decisions during intersection handling.

\subsubsection{CARLA Failure Cases}
We highlight two failure cases on the CARLA benchmark.

\begin{figure*}[htbp]
\centering
\includegraphics[width=1.0\linewidth]{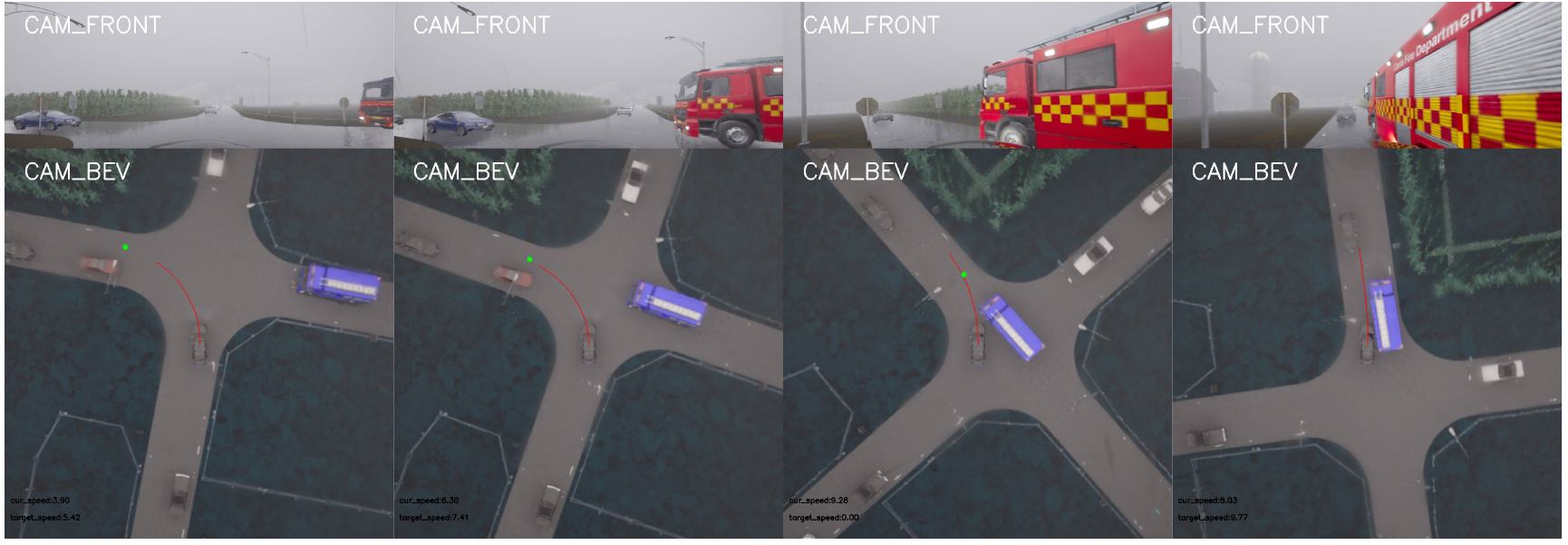}
\caption{Closed-loop failure case on CARLA. The model fails to yield to a potentially straight-driving vehicle, resulting in delayed braking and a collision.}

\label{fig:qual_7}
\end{figure*}

\paragraph{Case 1 – Failure to Yield to Oncoming Traffic.}  
In Figure~\ref{fig:qual_7}, the ego vehicle makes a left turn at an intersection but fails to yield to a truck approaching from the right side. As a result, it reacts too late and collides with the truck. This case highlights the model's limitation in proactively handling interactions with dynamic agents in intersection scenarios.

\begin{figure*}[htbp]
\centering
\includegraphics[width=1.0\linewidth]{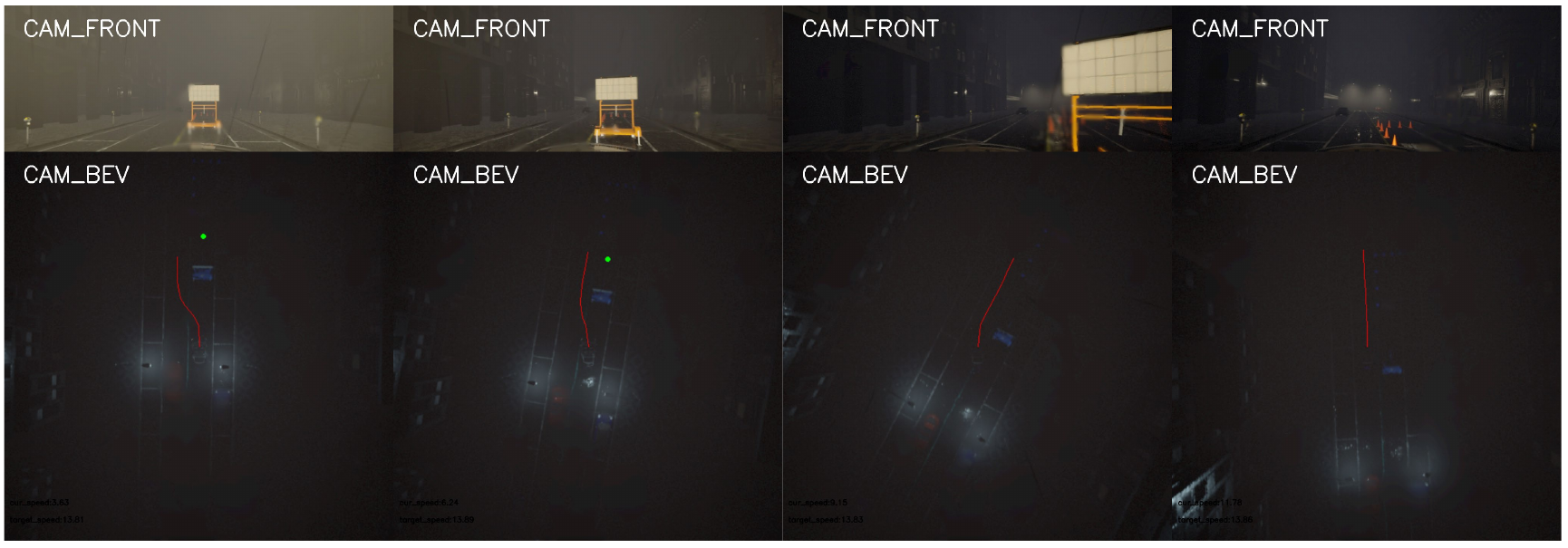}
\caption{Closed-loop failure case on CARLA. In a nighttime scenario, insufficient perception of fine details causes the ego vehicle to collide with an obstacle while attempting to avoid it.}

\label{fig:qual_8}
\end{figure*}

\paragraph{Case 2 – Obstacle Collision in Nighttime Scenario.}  
As shown in Figure~\ref{fig:qual_8}, under a low-light nighttime condition, the ego vehicle attempts to avoid a roadside obstacle. Due to insufficient perception of fine environmental details, the vehicle collides with the obstacle during the avoidance maneuver. This case illustrates the model’s vulnerability to degraded perception quality in challenging visual environments.

\clearpage

\end{document}